\documentclass[hidelinks, 12pt]{article}
\usepackage{manuscript}
\usepackage{xcolor}
\usepackage{enumerate}
\usepackage{booktabs,caption}
\usepackage[flushleft]{threeparttable}
\usepackage{pdflscape}
\usepackage{todonotes}
\usepackage{cancel}
\usepackage{rotating}
\usepackage{amsthm}
\usepackage{changepage}
\usepackage{array}
\usepackage{pgfplots}
\usepackage{tikz-dependency}
\usepackage{CJKutf8}

\hypersetup{
    colorlinks=true,
    linkcolor=blue,
    citecolor=blue,
    filecolor=blue,      
    urlcolor=blue,
    }

\newcounter{list-counter}

\title{Strategic resource allocation in memory encoding: An efficiency principle shaping language processing}

\author{Weijie Xu\thanks{Corresponding Author: \texttt{weijie.xu@uci.edu} (Weijie Xu)}\\
\and Richard Futrell\\
    }

\date{University of California, Irvine}

\begin{document}
{\setstretch{.8}
\maketitle
\begin{abstract}

\noindent
How is the limited capacity of working memory efficiently used to support human linguistic behaviors? In this paper, we propose Strategic Resource Allocation (SRA) as an efficiency principle for memory encoding in sentence processing. The idea is that working memory resources are dynamically and strategically allocated to prioritize novel and unexpected information. From a resource-rational perspective, we argue that SRA is the principled solution to a computational problem posed by two functional assumptions about working memory, namely its limited capacity and its noisy representation. Specifically, working memory needs to minimize the retrieval error of past inputs under the constraint of limited memory resources, an optimization problem whose solution is to allocate more resources to encode more surprising inputs with higher precision. One of the critical consequences of SRA is that surprising inputs are encoded with enhanced representations, and therefore are less susceptible to memory decay and interference. Empirically, through naturalistic corpus data, we find converging evidence for SRA in the context of dependency locality from both production and comprehension, where non-local dependencies with less predictable antecedents are associated with reduced locality effect. However, our results also reveal considerable cross-linguistic variability, suggesting the need for a closer examination of how SRA, as a domain-general memory efficiency principle, interacts with language-specific phrase structures. SRA highlights the critical role of representational uncertainty in understanding memory encoding. It also reimages the effects of surprisal and entropy on processing difficulty from the perspective of efficient memory encoding. \\


\noindent

\end{abstract}
}

\newpage


\section{Introduction}

Language processing in humans relies on working memory, a cognitive module known for its limited capacity to retain information \citep{baddeley1992working, just1992capacity, fedorenko2013direct}. Under this limitation, a linguistic signal, once perceived, is at the risk of being lost, rapidly overwhelmed by the continual torrent of new inputs \citep{christiansen2016now}. Meanwhile, language use seems effortless, with sophisticated linguistic representations being dynamically encoded and decoded within milliseconds. This dual nature of working memory raises the question: how is the limited capacity of working memory \textit{efficiently} used to support human linguistic behaviors?

In this paper, we propose \textit{Strategic Resource Allocation} (SRA) as an efficiency principle for memory encoding in sentence processing. Specifically, working memory resources are dynamically and strategically allocated to prioritize novel and unexpected information. We argue that this efficiency principle, as a resource-rational theory \citep{lieder2020resource, lewis2014computational, gershman2015computational}, naturally arises as the solution to a computational problem posed by two functional assumptions about working memory: its capacity is limited, and its representations are noisy. To examine this efficiency principle, we report three studies using naturalistic corpus data, where we demonstrate empirical support for strategic resource allocation through the lens of the locality effect in processing non-local syntactic dependencies.


\section{Strategic Resource Allocation (SRA)}\label{sec:background-SRA}

We first present the theoretical justification and existing empirical evidence for our proposal of Strategic Resource Allocation (SRA), drawing from the literature on sentence processing and psychophysics.

\subsection{Theoretical Proposal}

We propose that working memory resources are strategically allocated in a way that prioritizes novel and unexpected information given the context, an efficiency principle that we refer to as \textit{Strategic Resource Allocation} (SRA): 
\begin{enumerate}[(1)]
\setcounter{enumi}{\value{list-counter}}
    \item Strategic Resource Allocation (SRA) in memory encoding: \label{def:SRA}
    \begin{description}
        \item \textit{\textbf{Principle.}} Working memory resources are dynamically and strategically allocated in a way that prioritizes linguistic units that are unexpected and surprising given the context.
        \item \textit{\textbf{Core Prediction.}} The encoding of more surprising units is enhanced, resulting in more robust memory representations that are less susceptible to memory interference or decay.
    \end{description}
\setcounter{list-counter}{\value{enumi}}
\end{enumerate}
This principle is in line with the resource-rational analysis of human mind \citep{lieder2020resource, lewis2014computational, gershman2015computational}. Grounded in the bounded-rational approach to cognition \citep{anderson1990adaptive, simon1955behavioral}, resource-rational analysis aims to integrate the functional goals of a computational problem into the structural constraints of the underpinning cognitive architecture, providing a linkage between the computational-level and the algorithmic-level theories \citep{Marr:1982:VCI:1095712}. In other words, instead of looking for an unbounded optimization, resource-rational analysis seeks to explain human behaviors under bounded rationality, that is, to identify an optimal solution that strikes the balance between maximizing the functional utility and adhering to the structural constraints of the cognitive system.

In this section, we will first outline the computational problem faced by working memory: to infer past information from uncertainty with maximal accuracy under the constraint of limited memory resources. We will then explain how SRA provides an optimal solution to this computational problem.

\subsubsection{Inferring from uncertainty: A computational problem of working memory}

\begin{figure}
    \centering
    \includegraphics[width=0.5\linewidth]{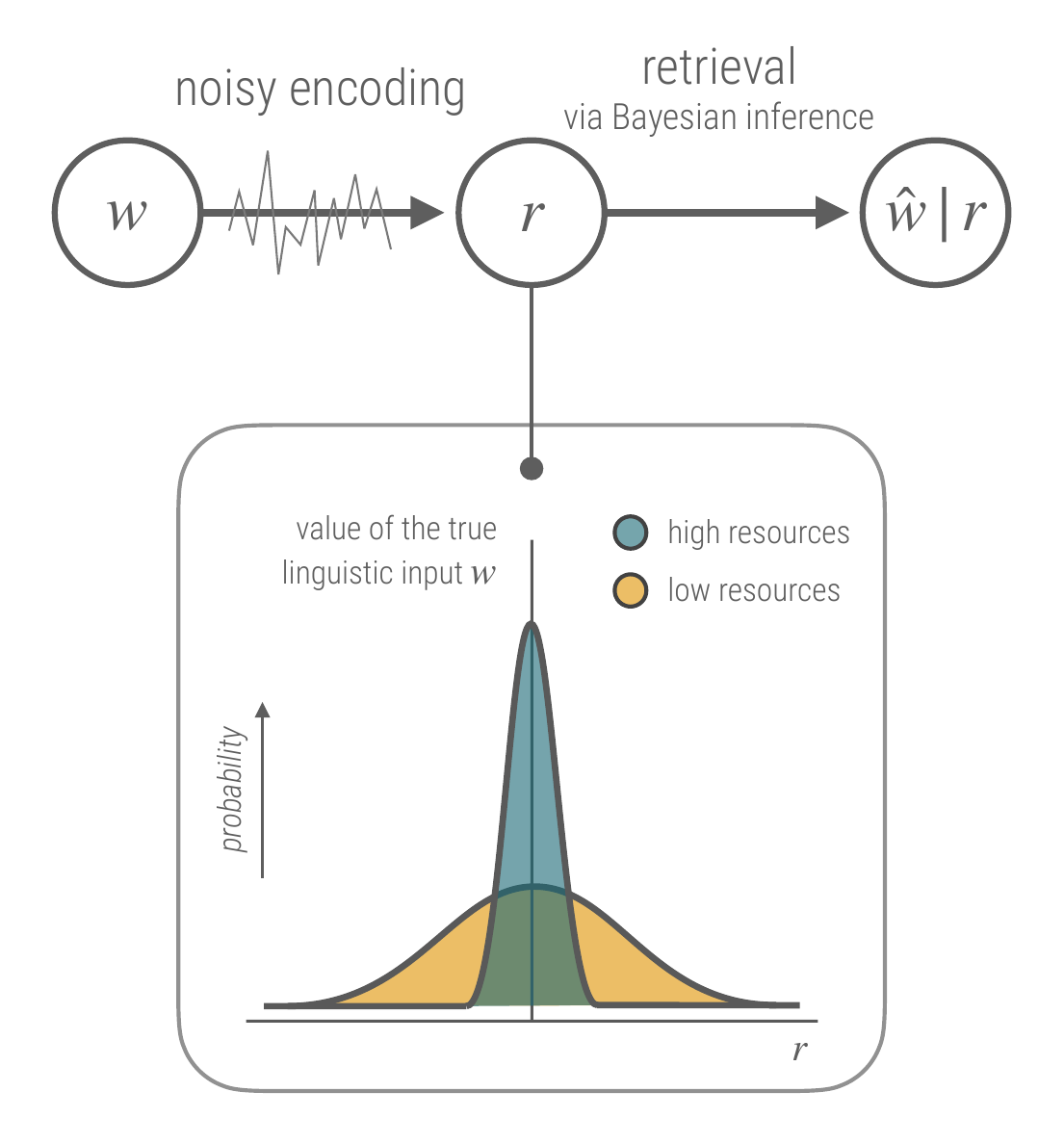}
    \caption{Working memory processes under the probabilistic framework. The linguistic input $w$ is encoded with noisy internal memory representation $r$. Higher memory resources results in sharper representation concentrated around the true input with less uncertainty. Memory retrieval can be considered an inference process reconstructing linguistic input based on noisy representation.}
    \label{fig:noisy-encode-demo}
\end{figure}

As already mentioned, the resource-rational explanation for SRA is rooted in a computation problem posed by two functional assumptions about working memory. First, the capacity of working memory is limited. Although the exact nature of this limitation is still under debate, recent models in some non-linguistic domains have shifted from a discrete slot representation \citep{miller1956magical, cowan2001magical, luck1997capacity, pashler1988familiarity} towards a continuous resource-based representation, where the limited resources can be flexibly allocated across the encoded information \citep{ma2014changing, van2012variability, brady2016working, brady2013probabilistic, van2018resource, sims2012ideal, sims2016rate, jakob2023rate, bates2020efficient}. 

\begin{figure}[t]
    \centering
    \includegraphics[width=1\linewidth]{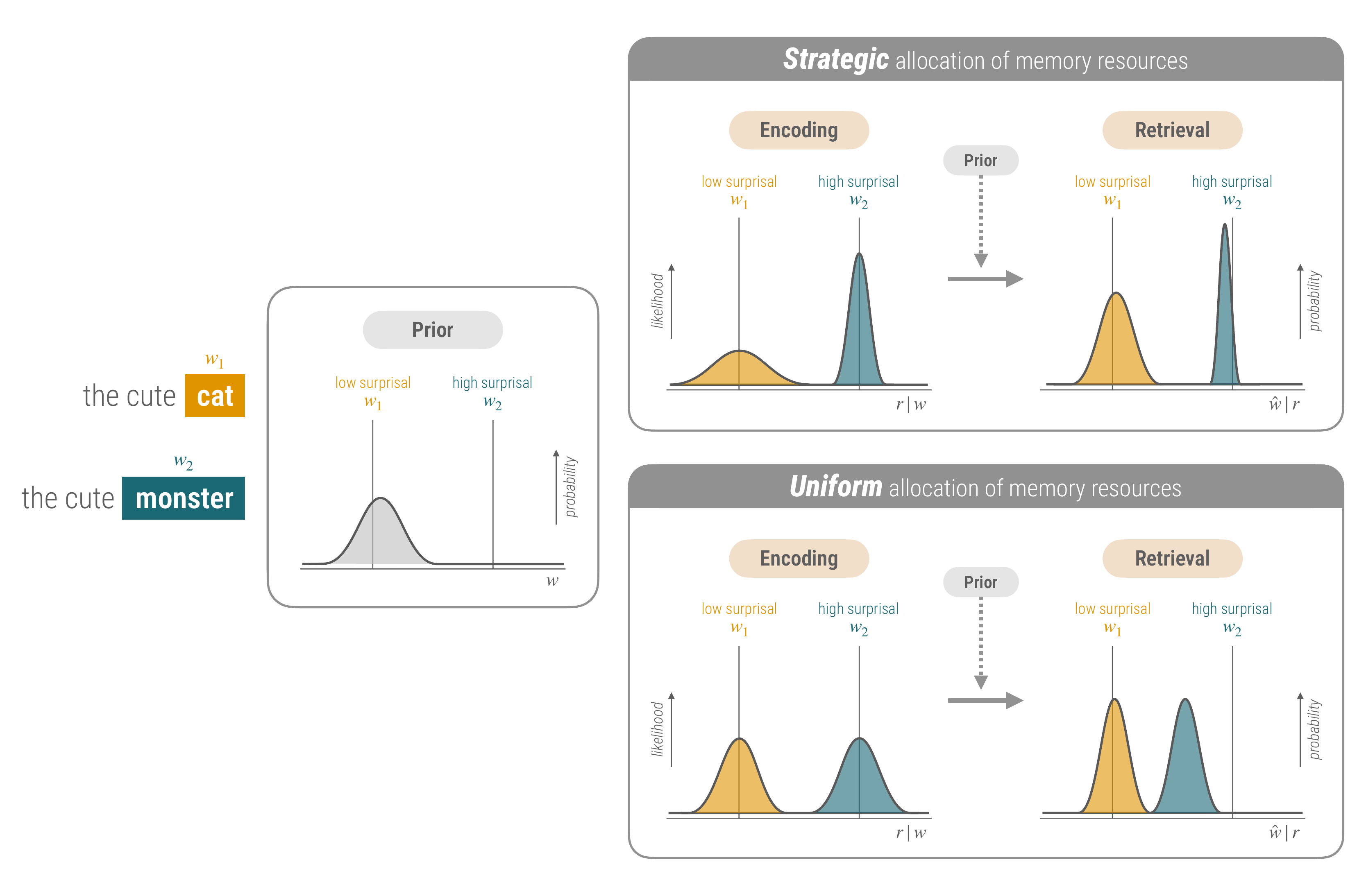}
    \caption{Conceptual framework of strategic resource allocation (SRA) versus uniform resource allocation in memory encoding. Compared to uniform allocation, SRA holds that more surprising units receive more memory resources, therefore less representational uncertainty in its encoding distribution. Upon retrieval, more surprising units are less reconstructable based on prior compared to less surprising ones. But SRA leads to lower retrieval error overall by improving the reconstruction of high-surprisal units with minimal loss to the retrieval accuracy on low-surprisal units.}
    \label{fig:SRA-demo}
\end{figure}

Second, memory representation is full of noise and uncertainty, with unpredictable corruption in the veridical forms of sensory input, resulting in distorted representations that undermine behavioral performance such as inaccurate recall and illusive comprehension \citep{ma2014changing, gibson2013rational, levy2008noisy, brady2024noisy, ferreira2002good}. This uncertainty is often represented under the probabilistic framework. As shown in Figure~\ref{fig:noisy-encode-demo}, when a linguistic input $w$ is received, it can be encoded into an internal representation through certain memory model $r = M(w)$. This $r$ is a probabilistic distribution centered around the true value of that input, such that the true input bears the highest probability in the encoding distribution compared to other alternatives.\footnote{In many psychophysics studies, the encoded representation is assumed to be a specific stimulus value that is generated from certain probabilistic distribution. Here in our work, we take a different assumption and postulate that what has been encoded, instead of a specific stimulus value, is the distribution itself, either through sampling \citep{hoover2023plausibility} or through probabilistic population codes \citep{ma2006bayesian}.} Given this noisy representation, the true state of a past input is inaccessible, and memory recall, decoding, or retrieval, is effectively an inferential process that reconstructs past input from uncertainty using the statistical structure of long-term knowledge. The results of this process are often mathematically characterized using Bayes' rule \citep[e.g.,][]{futrell2020lossy, gibson2013rational, levy2008noisy, ryskin2021erp, bays2024representation}:
\begin{equation} \label{eq:Bayes}
p(\hat{w} \mid r) \propto p_M(r \mid w)p(w).
\end{equation}
The equation describes a rational Bayesian decoder which infers the input from a specific memory representation $r$ integrating prior knowledge $p(w)$, yielding a posterior distribution $p(\hat{w} \mid r)$. Then, marginalizing over all possible values of $r$, the distribution on the reconstructed word given the true input $w^*$ is 
\begin{equation} \label{eq:Bayes}
p(\hat{w} \mid w^*) = \int p(\hat{w} \mid r)p(r \mid w^*) d r.
\end{equation}
In this inferential process, inputs that are more probable in the prior are more likely to be accurately reconstructed, resulting in higher retrieval accuracy. See Appendix~\ref{sec:retrieval-bayesian} for detailed mathematical formalization of this probabilistic memory encoding and retrieval process.

The two assumptions above naturally give rise to the following optimization challenge: \textit{how to maximize memory accuracy under the constraint of limited resources?} At the core of this challenge lies an \textit{efficiency} problem for two reasons. First, there is a functional goal, which is memory accuracy, against which the working memory performance is evaluated. Such a functional nature situates the current proposal under the rationalist approach to human mind. Second, working memory has internal constraints, in the sense that there is something it cannot achieve due to the cost from its own structure. Without such constraints, there would be no reason to look for an \textit{efficient} implementation of a functional goal. The acknowledgment of system-internal cost, therefore, further situates the current proposal under resource-rationality.\footnote{In this paper, we simply represent the internal cost of working memory as a computational bound (i.e., the total amount of available memory resources) within which a given task such as memory retrieval is imperfectly optimized, an approach that has been termed \textit{bounded optimality} \cite{icard2023resource}.} Next, we will explain how SRA provides a principled solution to the computational challenge outlined above.

\subsubsection{SRA as a resource-rational solution}

First of all, an important assumption we make is that the precision of the encoded distribution is proportional to the amount of memory resources allocated to an input unit. That is, as illustrated in Figure~\ref{fig:noisy-encode-demo}, more resources allocated to encode $w$ results in sharper distribution with less uncertainty, such that more probability mass is concentrated around the true input value $w$. Despite its lack of attention in the field of sentence processing, this assumption has been widely entertained in the literature of psychophysics \citep[e.g.,][]{bates2020efficient, bays2024representation, bays2009precision, ma2014changing}.\footnote{One of the biggest challenges to apply such an encoding distribution in the domain of language is the specification of hypothesis space. In psychophysics, the hypothesis space is usually a quantitative spectrum that can be objectively specified based on certain physical features. But for language, the linguistic inputs are discrete units. Nowadays, with the advance of modern NLP techniques, this challenge has been significantly mitigated given the distributive word representations such as word embeddings. However, it is still nontrivial work to figure out what kind of probabilistic distribution should be applied to word embedding space. See Appendix~\ref{sec:retrieval-bayesian} for a Gaussian approximation to the probabilistic memory processes.} As shown below, such a relationship between allocated resources and representational uncertainty lays the foundation for the derivation of SRA.

Before demonstrating the rationale behind SRA, let us first consider a naive strategy in which memory resources are uniformly distributed across all linguistic units regardless of the statistical structure in the context (Figure~\ref{fig:SRA-demo}, bottom panel). Under this uniform distribution, each input unit will receive an encoding distribution with identical precision. However, under the influence of prior, inputs that are more surprising under the prior will be less reconstructable, and the retrieval distribution will be more drawn towards the prior. Due to this difference in reconstructability, the same encoding distribution would yield different retrieval accuracy, disproportionally exerting impact on high surprisal inputs, reducing their retrieval accuracy more significantly than low surprisal ones. Therefore, a uniform distribution of memory resources is not the most efficient way to go for memory encoding, leaving substantial room for the improvement of overall retrieval accuracy.

Now, consider SRA, the strategic allocation of memory resources (Figure~\ref{fig:SRA-demo}, top panel). Recall that the idea is to strategically allocate more resources on linguistic units of higher surprisal \textit{a priori}. That means, the prioritized more surprising units will receive sharper encoding with higher precision. This asymmetric allocation of resources is more efficient than the uniform strategy described in the last paragraph, since it achieves higher accuracy on average across inputs. By sacrificing a slight reduction in retrieval accuracy for low surprisal units, significant gains can be achieved for high surprisal ones by preventing these irrecontructable units from being distorted in the first place. Put simply, when only a limited number of linguistic units can be encoded with minimal distortion, it is more important to encode the more surprising and less reconstructable ones (See Appendix~\ref{sec:minimize-error} for mathematical derivation).\footnote{By having an encoding strategy that optimizes faithful reconstruction of the true input, an implicit assumption we made is that the input signal is considered error-free. If the input itself contains errors (e.g., when there are speech errors produced by the speaker), a reconstruction that is strongly influenced by the prior may actually be preferred so that the signal errors can be corrected. How to deal with the errors in input signals is an important online processing task. But in the current proposal, we choose to analyze working memory as a system of information storage, whose main goal is to accurately encode and decode the information it receives \citep[cf.][]{hasson2015hierarchical}.}

Beyond memory, SRA aligns with theories such as predictive coding, free energy principle, and implicit learning. At the neural level, the predictive coding mechanism \citep{murray2002shape, rao1999predictive, aitchison2017or, blank2016prediction, sohoglu2020rapid, sohoglu2016perceptual, gagnepain2012temporal} and the free energy principle \citep{friston2005theory, friston2010free, gershman2019does} hold that the brain seeks to minimize its prediction error, or surprise, as a way to optimize its internal model of the external environment. This principle is implemented by encoding prediction errors rather than the raw sensory input in neural signals. At the behavioral level, implicit learning theories often hold that learning is error-driven, with considerable empirical support showing that larger prediction errors lead to greater learning effect \citep{chang2006becoming, scheepers2003syntactic, hartsuiker1998syntactic, bock1986syntactic, ferreira2003persistence, jaeger2013alignment, xu2024hierarchical, rumelhart1986learning, elman1990finding, courville2006bayesian,wagner1972inhibition}. Taken together, all these theories delivered a similar implication for our proposal in the domain of working memory. That is, when predictions conflict with the actual perceptual input (that is, when there is high surprisal input), it signals the need for comprehenders to update their mental model in order to make more accurate predictions in the future. Given this critical role of more surprising linguistic units in refining the mental model, it is reasonable to allocate more memory resources to them.

\subsection{Predictability-Precision Trade-Off}

\begin{figure}
    \centering
    \includegraphics[width=0.45\linewidth]{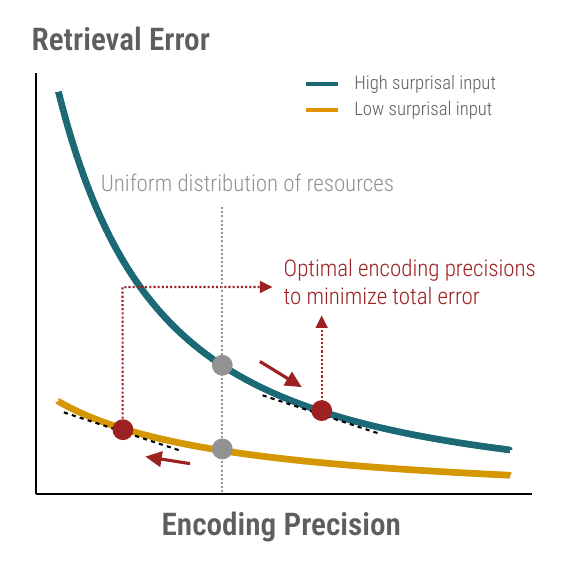}
    \caption{Retrieval error as a function of encoding precision for high-surprisal and low-surprisal inputs. The optimal encoding strategy balances the potential gain and loss in retrieval accuracy across linguistic inputs.}
    \label{fig:trade-off-demo}
\end{figure}

SRA can also be construed as a \textit{predictability-precision trade-off}, as illustrated in Figure~\ref{fig:trade-off-demo}. Imagine there are multiple linguistic units to be stored in working memory, with the goal of minimizing their total retrieval error (or, maximizing the retrieval accuracy). In most cases, for each unit, higher memory resources suggest higher encoding precision, which lead to lower retrieval error.\footnote{In some borderline cases, where the input word is too close to the prior prediction and the prior precision is too unreliable, the retrieval error may not monotonically decrease with increasing encoding precision. However, as shown in Appendix~\ref{sec:minimize-error}, in either case, the strategic resource allocation should still hold in the sense that more resources should be allocated to high surprisal units in order to minimize the expected total error.} However, given a fixed amount of resources, allocating more to one unit necessarily reduces what is available to others. Therefore, the gain of retrieval accuracy for one unit necessarily comes with the loss for others. Consequently, to minimize total retrieval error, the \textit{optimal} distribution of a fixed amount of resources should balance the gain and the loss in retrieval accuracy across units.

Where, then, does the balance hold? The intuition is illustrated in Figure~\ref{fig:trade-off-demo}. First, let us start from the uniform distribution of resources, where both high- and low-surprisal inputs receive the same amount of resources, and are thus encoded with the same degree of precisions (i.e., gray dots in Figure~\ref{fig:trade-off-demo}). Importantly, the retrieval error decreases faster for high-surprisal input. Therefore, at this point of uniform distribution of resources, there is a momentum to redistribute more resources to high-surprisal input. This is because such a redistribution towards high-surprisal input reduces the error faster. In other words, given a fixed amount of resources, the pressure to lower the overall retrieval error will push the encoding precision of high-surprisal input to increase from a uniform distribution of resources, and vice versa for low-surprisal one. This redistribution continues until there is balanced marginal effects, that is, when the slopes of the two error functions are equal, as in Figure~\ref{fig:trade-off-demo} (see Appendix~\ref{sec:minimize-error} for details).

\subsection{Precision, Accuracy, and Robustness}

It is important to point out that the key prediction of SRA is not simply about the mean accuracy of memory retrieval. As shown in Figure~\ref{fig:SRA-demo} (top panel), both predictable and unpredictable units can achieve relatively high accuracy with respect to the mean of retrieval distribution. For predictable input, this is supported by prior knowledge; for unpredictable input, this is achieved by more precise encoding representation. In fact, one of the critical consequences of SRA is that the retrieval mean accuracy should remain approximately similar across different linguistic units.\footnote{SRA does not necessarily predict the retrieval mean accuracy to be an absolute constant across all linguistic units. In fact, the accuracy for low surprisal units may still be higher than high surprisal ones in Bayesian inference. However, due to SRA, this difference can be reduced compared to a naive encoding strategy such as uniform resource allocation.} 

But what \textit{does} differ is the \textit{precision}, or \textit{uncertainty} in the representation. By allocating more resources, more surprising linguistic units are encoded with lower uncertainty (see the sharper distributions with high resources in Figure~\ref{fig:SRA-demo}). We thus propose that the uncertainty in the memory representation, rather than being linked to retrieval accuracy, is more directly related to \textit{memory robustness}. That is, memory representation of higher robustness is less susceptible to the interference from other elements in memory. It of course remains a debatable question what the linking hypothesis is for representational uncertainty, but the linkage between uncertainty and robustness gives us a working hypothesis that is readily testable, as shown below in the rest of this paper. We will return to this point later in General Discussion \ref{sec:role-uncertainty}. 

Our theoretical framework of SRA alludes to three effects on memory encoding precision:
\begin{enumerate}[(1)] \label{SRA-corollaries}
\setcounter{enumi}{\value{list-counter}}
    \item Three predicted effects of strategic resource allocation on encoding precision \label{SRA-corollaries}
    \begin{enumerate}[a.]
        \item \label{coro:input-surp} \textit{Effect of input surprisal}\\
        Surprising linguistic units bear higher encoding precision, resulting in more robust memory representation against interference.
        \item \label{coro:target-acc} \textit{Effect of memory constraint}\\
        More available memory resources result in higher encoding precision overall.
        \item \label{coro:prior-prec} \textit{Effect of prior precision}\\
        Precision of prior prediction does not necessarily increase or decrease encoding precision.
    \end{enumerate}
\setcounter{list-counter}{\value{enumi}}
\end{enumerate}

The most important prediction of SRA is (\ref{coro:input-surp}), the effect of input surprisal. As outlined earlier, the optimal strategy to minimize overall retrieval error is to allocate more memory resources to encode surprising input. This strategy results in higher precision in the representation, thus higher memory robustness against interference. This is the critical prediction that we are going to examine empirically in the current study.

For the effect of memory constraint (\ref{coro:target-acc}), SRA predicts that more available memory resources results in higher encoding precision in general. This will also lead to higher memory robustness and more accurate retrieval overall.

For the precision of prior prediction (\ref{coro:prior-prec}), its effect on encoding precision is in fact less straightforward. When the true input is very close to the prior prediction, it is indeed possible that less uncertainty in the prior can better support memory retrieval, thus less precise encoding is needed. However, when the true input is far from the prior prediction, it is not necessarily the case that more precise prior can still support better retrieval. We will discuss this effect in more detail and its implication for the effect of prediction entropy on processing difficulty in General Discussion \ref{sec:encoding-diff}.

\subsection{Some Existing Empirical Evidence}

A dynamic similar to strategic resource allocation (SRA) is observed in the resource-rational model of sentence processing by \cite{hahn2022resource}. Grounded in the framework of lossy-context surprisal \citep{futrell2020lossy}, their model involves a contextual representation that represents only those words that are most useful for a downstream next-word prediction task. Their model predicts that function words, which are mostly predictable from the linguistic context, are more likely to undergo decay. In fact, our proposal of strategic resource allocation and lossy-context surprisal theory form two sides of the same coin in many aspects. We will discuss the relationship between these two theories in General Discussion Section~\ref{sec:lossy-context}.

Studies focusing on memory retrieval mechanisms find that linguistic units of higher semantic complexity can be more easily retrieved in later stages of processing despite the initial encoding difficulty, implicating an enhanced accessibility for informative content from the model-theoretic perspective of informativity \citep{hofmeister2011representational, hofmeister2014distinctiveness, karimi2016informativity, troyer2016elaboration}. However, the exact cognitive underpinning for this empirical observation is still debatable \citep{hofmeister2014distinctiveness, karimi2023delayed}, and there is lack of clear empirical evidence for whether this effect can be extended to the information-theoretic view of informativity based on probabilistic prediction \citep{shannon1948mathematical}. In spite of these unsettled issues, as a preliminary evidence from the existing literature, the effect of facilitated retrieval for semantically complex units aligns with our rational account outlined above, in the sense that the enhanced accessibility associated with informative units results from the prioritized resources allocated to their encoding.

Recently, SRA is more directly examined by \citet{xu2025informativity} through the lens of the agreement attraction effect in English. As shown below, even though the sentences in (\ref{ex:AgreeAttr}) are ungrammatical in English due to the mismatch of number feature between the subject head noun and the main verb, they are often perceived grammatical by native speakers due to the interference from the distractor noun in between, which shares the number feature with the ungrammatical main verb. In \cite{xu2025informativity}, by manipulating the surprisal of the subject head noun through a prenominal adjective, they find that, compared to more surprising subject head nouns (e.g., cute \textit{monster}), less surprising ones (e.g., evil \textit{monster}) lead to stronger agreement attraction effect, such that the processing of the main verb is less susceptible to the interference from the distractor noun. They interpret the result as evidence for an enhanced memory representation of more surprising linguistic units against memory interference.
\begin{enumerate}[(1)]
\setcounter{enumi}{\value{list-counter}}
    \item \label{ex:AgreeAttr}
    \begin{enumerate}[a.]
        \item \label{ex:AgreeAttr-pred} *The evil \textit{monster} who chased the kids seemingly \textit{were} gone before the sunset. \hfill [low surprisal]
        \item \label{ex:AgreeAttr-unpred} *The cute \textit{monster} who chased the kids seemingly \textit{were} gone before the sunset. \hfill [high surprisal]
    \end{enumerate}
\setcounter{list-counter}{\value{enumi}}
\end{enumerate}

In visual working memory, statistical regularities in long-term knowledge have been shown to shape memory performance. Despite the fact that items more consistent with prior knowledge are easier to be encoded with lower neural activity and enhanced behavioral performance \citep{bates2020efficient, blalock2015stimulus, jackson2008familiarity, xie2017familiarity, girshick2011cardinal}, some recent studies indeed observe that, in later stages of processing, these familiar items are de-prioritized to save more resources for the processing of novel ones \citep{kowialiewski2022between, bruning2020long, hedayati2022model, brady2024noisy}. For example, in a delayed-estimation task, \cite{bruning2020long} ask participants to first memorize and then recall the exact locations of six colored balls on a circle after a brief delay. Before the task, a sub-area on the circle has been previously illustrated to certainly contain the ball with a specific color (e.g., red ball) as a prior information. Their critical finding is that colors not included in the prior information (e.g., non-red balls) have lower recall accuracy when positioned closer to that sub-area, suggesting that memory resources have been shifted away from the prior area to prioritize other areas where novel information is more likely to appear.

\section{SRA and Dependency Locality}

The empirical focus of this paper to examine SRA is the \textit{locality effect} in sentence processing, which has been considered a representative example of the efficient use of working memory resources. In this section, we will first introduce the empirical background of dependency locality effect. Then, we will present the empirical predictions of SRA in the context of dependency locality.

\subsection{Dependency Locality}

Consider the sentence pair in~(\ref{ex:DLT}). In (\ref{ex:DLT-local}), codependents in the subject--verb dependency are adjacent to each other, whereas in (\ref{ex:DLT-nonlocal}), there is additional linguistic material in between:
\begin{enumerate}[(1)]
\setcounter{enumi}{\value{list-counter}}
    \item \label{ex:DLT}
    \begin{enumerate}[a.]
        \item \label{ex:DLT-local} The \textit{monster} \underline{approached} the princess...
        \item \label{ex:DLT-nonlocal} The \textit{monster} who stayed in the tower \underline{approached} the princess...
    \end{enumerate}
\setcounter{list-counter}{\value{enumi}}
\end{enumerate}
The Dependency Locality Theory (DLT) \citep{gibson1998linguistic, gibson2000dependency} holds that the formation of the non-local structures is constrained by the limited capacity of working memory. Specifically, as dependency distance increases, there is a higher memory cost to store the incomplete dependency as well as a higher integration cost to compute the new structural representation when the other codependent is encountered. In support of DLT, increased processing difficulty is often associated with structures that have longer dependency distance \citep[e.g.,][]{miller1964free, yngve1960model, grodner2005consequences, bartek2011search, ford1983method, king1991individual, gordon2001memory, traxler2002processing}.\footnote{There is actually an anti-locality effect often found in some head-final dependencies, which is considered to be better explained by an expectation-based mechanism \citep{vasishth2006argument, nakatani2010line, konieczny2000locality, levy2013expectation}.} Similarly, in the resolution of structural ambiguity where a constituent has multiple potential attachment sites, there is a tendency for comprehenders to prefer the structure with local attachment \citep{frazier1978sausage, pearlmutter2001recency, gibson1996recency}. 

Due to the memory constraint involved in processing non-local dependencies, an efficiency principle for language structure should be that linguistic units connected in a syntactic dependency tend to stay close in linear order. This locality principle is evidenced by cross-linguistic word-order patterns \citep{hawkins1990parsing,hawkins1994performance,hawkins2004efficiency,ferrericancho2004euclidean,liu2008dependency,futrell2015large,futrell2020dependency,temperley2018minimizing,liu2021crosslinguistic,liu2023development} \citep[cf.][]{liu2020mixed}, and has been argued to explain typological patterns such as the consistency in head direction, the contiguity of constituents, and the asymmetry of short-before-long versus long-before-short between head-initial and head-final languages \citep{futrell2020dependency,hawkins2004efficiency, hawkins1994performance}. 

More recently, some studies propose a generalization from dependency locality to information locality, where any pair of linguistic units with high co-occurrence statistics, no matter whether they are in the same syntactic dependency or not, should stay close in linear order \citep{futrell2019information,futrell2020lossy, hahn2021modeling, hahn2022crosslinguistic}. Compared to previous work, these studies highlight the role of predictive processing, pointing out an interaction between the memory-based and the expectation-based mechanisms. Specifically, under the framework of Surprisal Theory \citep{levy2008expectation, hale2001probabilistic}, the processing difficulty of a linguistic unit is proportional to how well it is predictable from the memory representation of the past input, which is prone to memory loss and distortion. The locality effect, as an efficient use of working memory, suggests that linguistic units carrying the most relevant information to predict the current one should stay in the recent past before they are forgotten.

These locality principles depend on the precise nature of working memory. Therefore, beyond the general capacity-based constraint proposed by DLT, it remains an open question how far this efficiency account can go with more and more realistic and detailed characterization of the nature of working memory constraints. Moreover, the existing discussion in the literature rarely addresses efficiency in processing per se. In other words, it is possible that memory limitations make language users not only actively choose a sentence form that is easier to process, but also develop an efficient processing strategy to better handle the information they passively receive.

\subsection{The Current Study}

\begin{figure}[htp]
    \centering
    \includegraphics[width=0.8\linewidth]{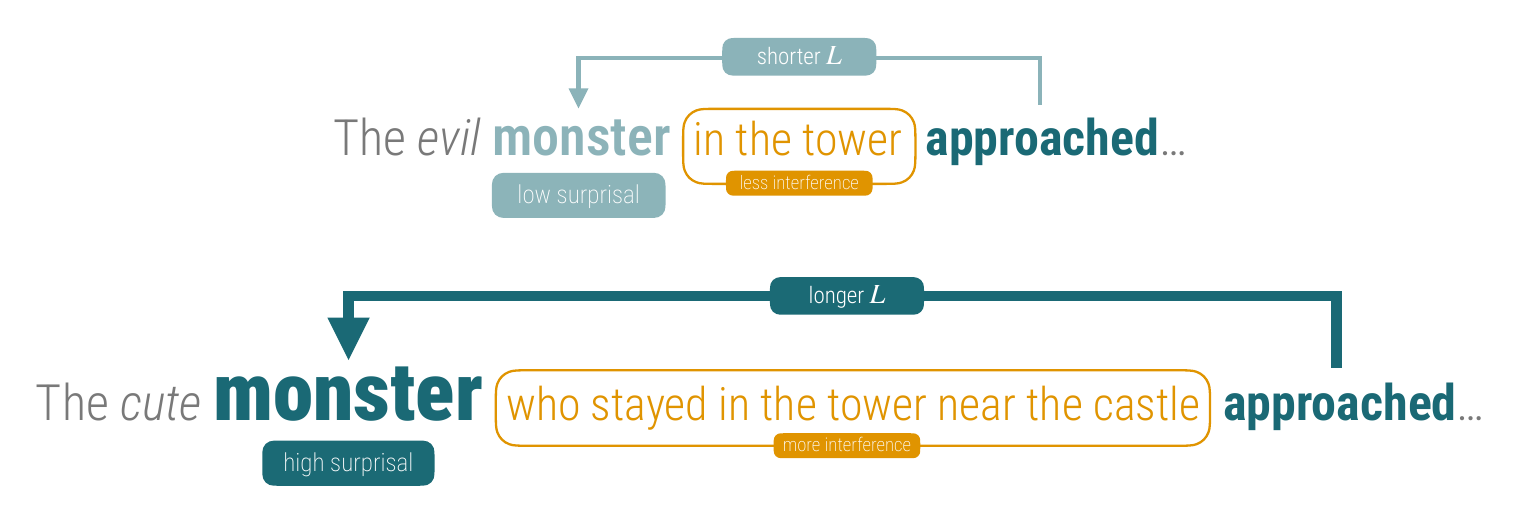}
    \caption{Empirical prediction of strategic resource allocation in dependency locality. High-surprisal antecedents are more tolerable to longer dependency length.}
    \label{fig:MemDepend-demo}
\end{figure}

We examine SRA in the context of dependency locality through naturalistic corpus data. If working memory resources are indeed dynamically and strategically allocated such that novel and unexpected information is prioritized, we predict that antecedents (i.e., left codependents) that are more surprising should receive sharper encoding with less uncertainty. The consequence of this is that memory for more surprising antecedents is enhanced, making their representations less susceptible to memory decay and interference before they need to be re-accessed at the other side of the dependency. Therefore, as illustrated in Figure~\ref{fig:MemDepend-demo}, more surprising antecedents should be able to tolerate longer dependency length, resulting in a reduced locality effect. We approach this prediction from both production (Study~1) and comprehension (Study~2a and 2b). 

There are two terminological clarifications. First, we adopt a relatively broad interpretation of the term \textit{memory encoding} in this article, focusing on the representational aspect of working memory mechanisms. Second, the term \textit{resources} refers to any quantity that is limited and costly to use for better cognitive performance. Given the ongoing debate about the exact nature of working memory resources \citep{ma2014changing, bays2024representation}, we choose to restrict the use of this term to its abstract sense.

To preview our results, we find converging evidence from both production and comprehension that unexpected information is encoded with enhanced robustness against decay and interference. In Study 1, which focuses on production data, we observe that more surprising antecedents are associated with longer dependency lengths, an effect that is not reducible to a simple frequency effect. Moreover, the effect mostly exists within Indo-European and head-initial languages in our analysis, and is more consistent for subject relations. We discuss the cross-linguistic variability in General Discussion. In Study~2a and 2b, examining comprehension data from English reading-time corpora, we find a reduced locality effect at the retrieval site for more surprising antecedents. Consistent with Study~1, this effect is more pronounced in subject relations and is observed more reliably in the self-paced reading corpus (Study~2a) than in the eye-tracking corpus (Study~2b).

\section{Study 1: Production Side}

We first examine strategic resource allocation in dependency locality in production. We predict that in production, the pressure to minimize dependency length can be relaxed when the antecedent contains novel and unexpected information. Consider the subject--verb dependency in the sentences below in (\ref{ex:DL-SRA}):
\begin{enumerate}[(1)]
\setcounter{enumi}{\value{list-counter}}
    \item \label{ex:DL-SRA}
    \begin{enumerate}[a.]
        \item \label{ex:DL-SRA-pred} The evil \textit{monster} in the tower \underline{approached}...
        \item \label{ex:DL-SRA-unpred} The cute \textit{monster} who stayed in the tower near the castle \underline{approached}...
    \end{enumerate}
\setcounter{list-counter}{\value{enumi}}
\end{enumerate}
The subject ``the cute monster'' in (\ref{ex:DL-SRA-unpred}) is more surprising compared to ``the evil monster'' in (\ref{ex:DL-SRA-pred}). According to our hypothesis, the unpredictable ``cute monster'' should be prioritized with more memory resources for encoding, and therefore is more capable of resisting the interference or decay introduced by the intervening material before the verb. As a result, compared to (\ref{ex:DL-SRA-pred}), the less predictable antecedent in (\ref{ex:DL-SRA-unpred}) is able to tolerate more intervening material before being re-accessed at the retrieval site (i.e., the right codependent), leading to longer dependency length. We measure the predictability of word $w$ at position $t$ as surprisal $S_t$:
\begin{equation}
    \label{eq:surprisal-def}
    S_t \equiv -\log p\left(w_t \mid w_{<t}\right),
\end{equation}
which is the negative log likelihood of the word $w_t$ given its preceding context $w_{<t}$. The higher the surprisal, the less predictable a word is. Therefore, we predict a positive correlation between \textsc{antecedent surprisal} and dependency length $L$ in production data.

Besides \textsc{antecedent surprisal}, we also examined the role of \textsc{antecedent frequency} in shaping memory allocation. On the one hand, these two quantities are highly correlated, in that low frequency words are also unpredictable in general, thus yielding higher surprisal. However, on the other hand, compared to surprisal, frequency as a unigram probability does not contain any information from the context. By comparing the effects of \textsc{antecedent surprisal} and \textsc{antecedent frequency}, we aim to look into to what extent the contextual information contributes to the efficiency strategy of working memory encoding. We expect less frequent antecedents to associate with longer dependency length.

\subsection{Method}

\subsubsection{Data}

We used the corpora of 10 languages taken from Universal Dependencies (UD) release 2.11 \citep{nivre-etal-2020-universal}, as summarized in Table~\ref{tab:dep-datasets}, with the aim to cover a wide variety of typological configurations (e.g., head-initial vs. head-final; free vs. rigid word order).\footnote{The original Russian corpus has over 1.2M tokens with over 600 documents; we randomly sampled 300 documents from the original corpus in our analysis in order to save on computational power.} An illustration of UD annotations is shown in (\ref{ex:UD-annotation}), where each arc represents a dependency whose direction is from the head to the dependent.\footnote{In our analysis, antecedent is defined as the left codependent of a dependency, and the retrieval site is always considered the right codependent, although as seen in (\ref{ex:UD-annotation}) the direction of a dependency can either go from the left codependent to the right or the other way around.}

\begin{enumerate}[(1)]
\setcounter{enumi}{\value{list-counter}}
    \item \label{ex:UD-annotation} Example of UD annotation: 
    \begin{center}
    \begin{dependency}[arc edge, arc angle=85, text only label, label style={above}]
    \begin{deptext}[column sep=0.8em]
    Isaac \& Newton \& left \& a \& note \& to \& Einstein. \\
    \end{deptext}
    \depedge{3}{1}{nsubj}
    \depedge{1}{2}{flat}
    \depedge{3}{5}{obj}
    \depedge[arc angle=60]{5}{4}{det}
    \depedge{3}{7}{obl}
    \depedge{7}{6}{case}
    \deproot[edge unit distance=3.2ex]{3}{root}
    \end{dependency}
    \end{center}
\setcounter{list-counter}{\value{enumi}}
\end{enumerate}

Compared to the Surface Syntactic Universal Dependencies (SUD) \citep{gerdes2018sud}, which is another major project of dependency corpora, the UD annotation scheme is content-word-oriented. That is, UD always labels content words as the head of a unit. As a result, UD favors lexical heads rather than functional heads in cases like adpositions, subordinating conjunctions, auxiliaries, and copulas. For the sentence above in (\ref{ex:UD-annotation}), for example, SUD annotates the oblique relation as from the head ``left'' to the preposition ``to'' rather than to ``Einstein.'' In the current work, we chose to use UD corpora since memory processes are more sensitive to content words rather than function words \citep{gibson1998linguistic, grodner2005consequences}. 

Some UD corpora consist of out-of-context independent sentences, while others are organized document by document, which provide longer and enriched discourse context for each token. This difference may influence our surprisal estimates, which are sensitive to the preceding context of each token. We extracted all the dependencies of a sentence annotated in the UD corpora. All the UD corpora we used have a pre-defined split into training, dev, and test sets. We only used the pre-defined training sets since they already have decent sample size.

\begin{table}[]
\begin{adjustwidth}{-0.2in}{-0.2in}
    \centering
    \begin{tabular}{llcrrr}
        \toprule
         \textbf{Language} & \multicolumn{1}{c}{\textbf{Corpus}} & \textbf{Genre} & \textbf{\# All} & \textbf{\# Subj} & \textbf{\# Obj} \\
        \midrule
         Danish & DDT \citep{johannsen2015universal} & sent & 45,976 & 4,203 & 3,963 \\
         English & GUM \citep{zeldes2017gum} & doc & 89,947 & 7,881 & 7,296 \\
         German & GSD \citep{mcdonald-etal-2013-universal} & sent & 155,480 & 9,602 & 8,474 \\
         Italian & ISDT \citep{bosco2013converting} & doc & 208,939 & 10,323 & 11,735 \\
         Japanese & GSD \citep{tanaka-etal-2016-universal} & sent & 113,771 & 5,005 & 4,018 \\
         Korean & Kaist \citep{chun2018building} & doc & 154,609 & 9,855 & 24,690 \\
         Mandarin & GSDSimp \citep{nivre-etal-2020-universal} & sent & 63,456 & 5,538 & 7,576 \\
         Russian & SynTagRus \citep{droganova2018data} & doc & 329,745 & 32,822 & 25,065 \\
         Spanish & AnCora \citep{taule2008ancora} & doc & 333,728 & 21,472 & 31,143 \\
         Turkish & BOUN \citep{marcsan2022enhancements} & sent & 45,914 & 3,861 & 4,680 \\
         \bottomrule
         \end{tabular}
    \caption{Dependency corpora used in Study~1. `Genre' refers to whether the texts in the corpus are organized as independent sentences (`sent'), or as documents with larger coherent discourse size (`doc'). `\#~All' indicates the number of all the dependencies after data exclusion. `\#~Subj' is a subset of `\#~All' and indicates the number of dependencies with subject relations. `\#~Obj' indicates the number of dependencies with object relations.}
    \label{tab:dep-datasets}
\end{adjustwidth}
\end{table}

\subsubsection{Estimating Token Surprisal}

In this work, to ensure that the results are not the artifact of a specific language model, we generated surprisal measures from both the GPT-3 base \citep[\texttt{text-davinci-001}; ][]{brown2020language} and the mGPT language models \citep{shliazhko2024mgpt}, both being trained on multilingual data. For each token $w$ in the dependency corpora, we obtained its surprisal $-\log p(w_t \mid w_{<t})$ given the preceding context from both models. We used the maximally allowed context window in the corresponding document or sentence. It is worth noting that Mandarin Chinese, unfortunately, is not supported by mGPT. Therefore, we only report the results with GPT-3 surprisal for Mandarin.

Contemporary large language models (LLMs) implemented with artificial neural networks provide state-of-the-art probabilistic measures of linguistic sequences and next-word predictions for the approximation of human predictive processing in psycholinguistics research \citep{wilcox2020predictive,wilcox2023testing,xu2023linearity,shain2024large}. Empirically, the surprisal generated from LLMs highly correlates with human language processing difficulty indexed by both behavioral and neural responses \citep{goodkind2018predictive, hao2020probabilistic, shain2024large, hoover2023plausibility, schrimpf2021neural, li2023heuristic, wilcox2023testing, xu2023linearity, hu-etal-2020-systematic}. 

\subsubsection{Measuring Dependency Length}

We did the analysis with two different measures of dependency length $L$. The first measure is an orthographic one $L_\mathrm{O}$, which is the number of words between the codependents of a dependency. The second measure is an information-theoretic one $L_\mathrm{I}$, which sums up the surprisal of all words between codependents from $w_i$ to $w_{i+N}$: 
\begin{equation} \label{eq:info-length}
\begin{split}
L_\mathrm{I} &= -\log p(w_{i \dots i+N} \mid w_{<i}) \\
  &= -\sum^{i+N}_{j=i} \log p(w_{j} \mid w_{<j}).
\end{split}
\end{equation}
We used these two measures because different words presumably induce memory interference to different extents. For example, compared to a content word that marks a discourse referent, a function word such as a determiner is way less informative, and may require much smaller memory load, thus inducing weaker memory interference \citep{gibson1998linguistic, grodner2005consequences}. Compared to the orthographic $L_\mathrm{O}$, which treats all the words in the same way, the information-theoretic $L_\mathrm{I}$ may better capture the above-mentioned variability across different words \citep{hahn2021modeling}.\footnote{A potential problem with the information-theoretic $L_\mathrm{I}$ lies in the dual role attributed to surprisal: it has been theorized as being proportional both to the allocated memory resources and to memory cost. Although intuitively more memory resources allocated may induce higher memory cost as well, we acknowledge that the extent to which these two concepts can be treated as interchangeable remains a debatable question.}

\subsubsection{Data Transformation and Exclusion}

Constructions such as foreign phrases, multi-word proper names, and fixed expressions are annotated as flat structures in UD corpora. We merged flat structures such that the surprisal of the whole structure is the sum of all its components, and that the first word in the flat structure is treated as the head when calculating the length of a dependency. For example, the subject--verb dependency in (\ref{ex:UD-annotation}) involves a flat structure in the subject position. The antecedent surprisal for this dependency is thus the sum of surprisal over both words ``Isaac Newton,'' and the dependency length by word counts is 1, since the first word ``Isaac'' is one word away from the verb. We excluded sentences that are less than five-word long, since sentences that are too short may have limited room for the dependency length to vary and many of the short ``sentences'' are in fact titles and extended proper names (e.g., e-mail addresses and institution names). We excluded punctuation tokens. We also excluded tokens whose surprisal value is greater than 20 bits, as the surprisal estimates for such rare word sequences may be unreliable. Moreover, exceedingly surprising information may introduce confounding factors in human processing. We then extracted all the dependencies in which both the head and the dependent are spared from data exclusion.

\subsubsection{Data Analysis}

For each language, the analysis consists of three parts. The first one is on the full dataset obtained as introduced above, with all types of dependency relations included. In addition, we also took a closer look into the dependencies whose dependent is a core argument in the sentence. Therefore, we also ran analysis on two subsets of the full dataset above, which include subject relations\footnote{Annotated as \texttt{nsubj} and \texttt{csubj} in UD corpora.} and object relations\footnote{Annotated as \texttt{obj}, \texttt{iobj}, \texttt{ccomp}, and \texttt{xcomp} in UD corpora.} respectively.

For the analysis with the full dataset, for each language, we ran separate linear mixed-effects models predicting the two variants of dependency length $L$ as the dependent variable, using the \texttt{lmerTest} package in R \citep{kuznetsova2017lmertest}. The critical fixed-effect predictor is the \textsc{antecedent surprisal}, with random intercept by dependency types.\footnote{As mentioned above, we also compare the effect of \textsc{antecedent surprisal} with \textsc{antecedent frequency} in the current analysis. However, the models with random slopes for both effects rarely converge. Therefore, for better interpretability of the statistical result, we only included random intercept by dependency types.} For the analyses with subject and object relations, we ran linear models with the same fixed effects. We included five control variables for all the analyses, as in (\ref{control-vars-study1}). \textsc{Sentence position} aims to control the discourse-level information structure, where more information may be given as the discourse develops; \textsc{Antecedent position} aims to control that antecedents appearing towards to the end of a sentence naturally tend to have shorter dependency length. \textsc{Sentence length} aims to control for two possible confounds: first, longer sentences may tend to have longer dependency length in general; second, longer sentences may tend to have more complex syntactic structure, which may be associated with more surprising antecedents. We also included \textsc{antecedent frequency} in log scale retrieved from \cite{wordfreq} in order to see whether the surprisal effect is reducible to a simple frequency effect. For the analysis with information-theoretic dependency length $L_\mathrm{I}$, we included an additional control variable \textsc{baseline surprisal}, which is the surprisal averaged across all words within a sentence. This is to address the confound that sentences with higher baseline surprisal naturally leads to a positive correlation between antecedent surprisal and the information-theoretic $L_\mathrm{I}$. All variables are $z$-scaled.

\begin{enumerate}[(1)]
\setcounter{enumi}{\value{list-counter}}
    \item Control variables in Study~1 \label{control-vars-study1}
    \begin{itemize}
        \item \textsc{sentence position}: position of the sentence in the current document (\textit{only included if the corpus is organized document-by-document})
        \item \textsc{antecedent position}: position of the antecedent in the current sentence
        \item \textsc{sentence length}: length of the sentence measured as word counts
        \item \textsc{antecedent frequency}: log frequency of the left codependent
        \item \textsc{baseline surprisal}: average surprisal across all words within a sentence (\textit{only included for the analysis with information-theoretic dependency length $L_\mathrm{I}$})
    \end{itemize}
\setcounter{list-counter}{\value{enumi}}
\end{enumerate}

\subsection{Result}

\begin{figure}[]
    \centering
    \includegraphics[width=1\linewidth]{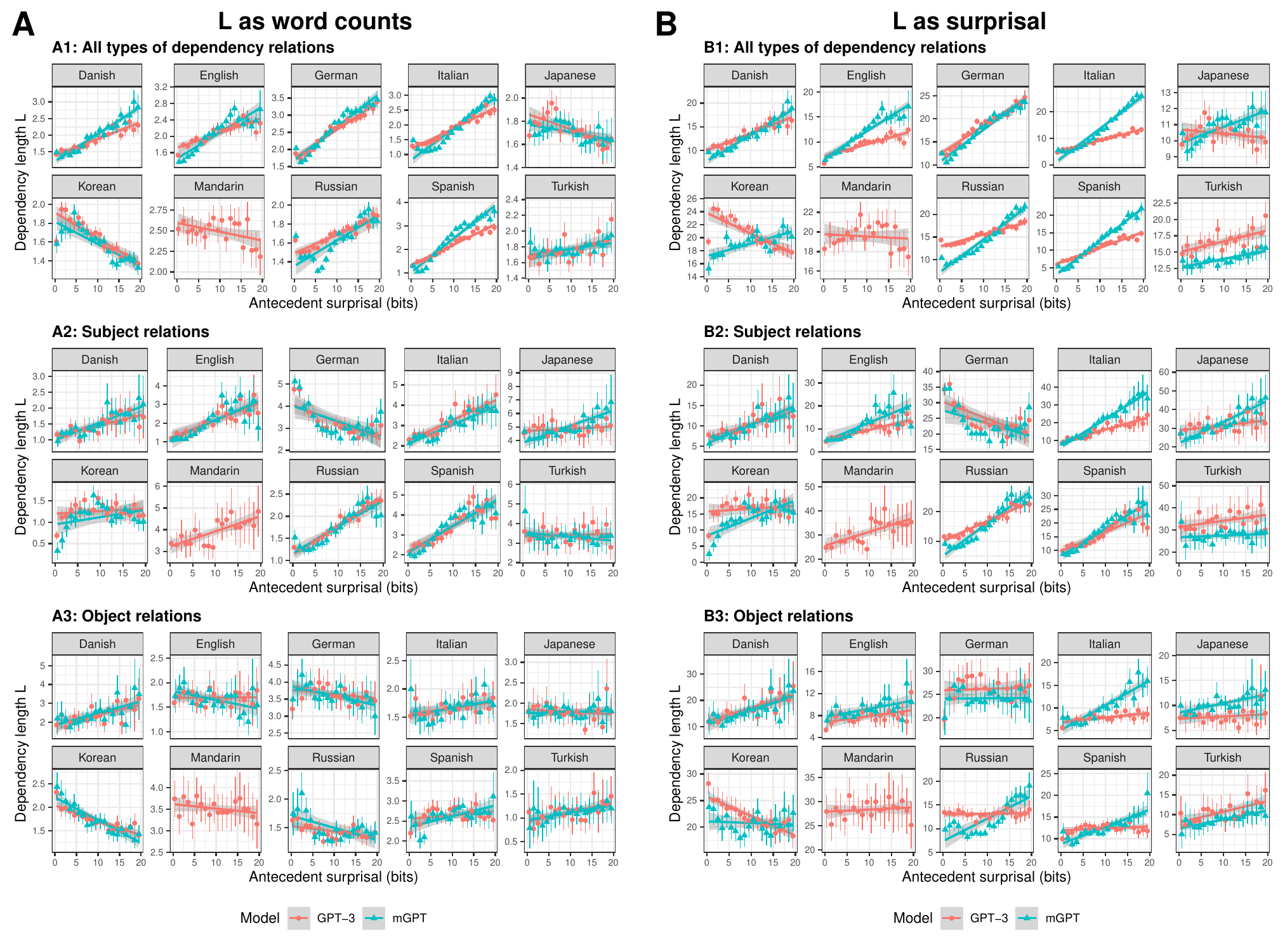}
    \caption{Dependency length $L$ as a function of \textsc{antecedent surprisal}. \textbf{\textit{Panel A}} corresponds to $L$ measured as intervening word counts. \textbf{\textit{Panel B}} corresponds to $L$ measured as the sum of surprisal over intervening words. Surprisal is binned into 20 categories, and the mean $L$ within each category is shown with a 95\% confidence interval. A linear fit to these points is presented.}
    \label{fig:Study-1-DL}
\end{figure}

\begin{figure}[]
    \centering
    \includegraphics[width=1\linewidth]{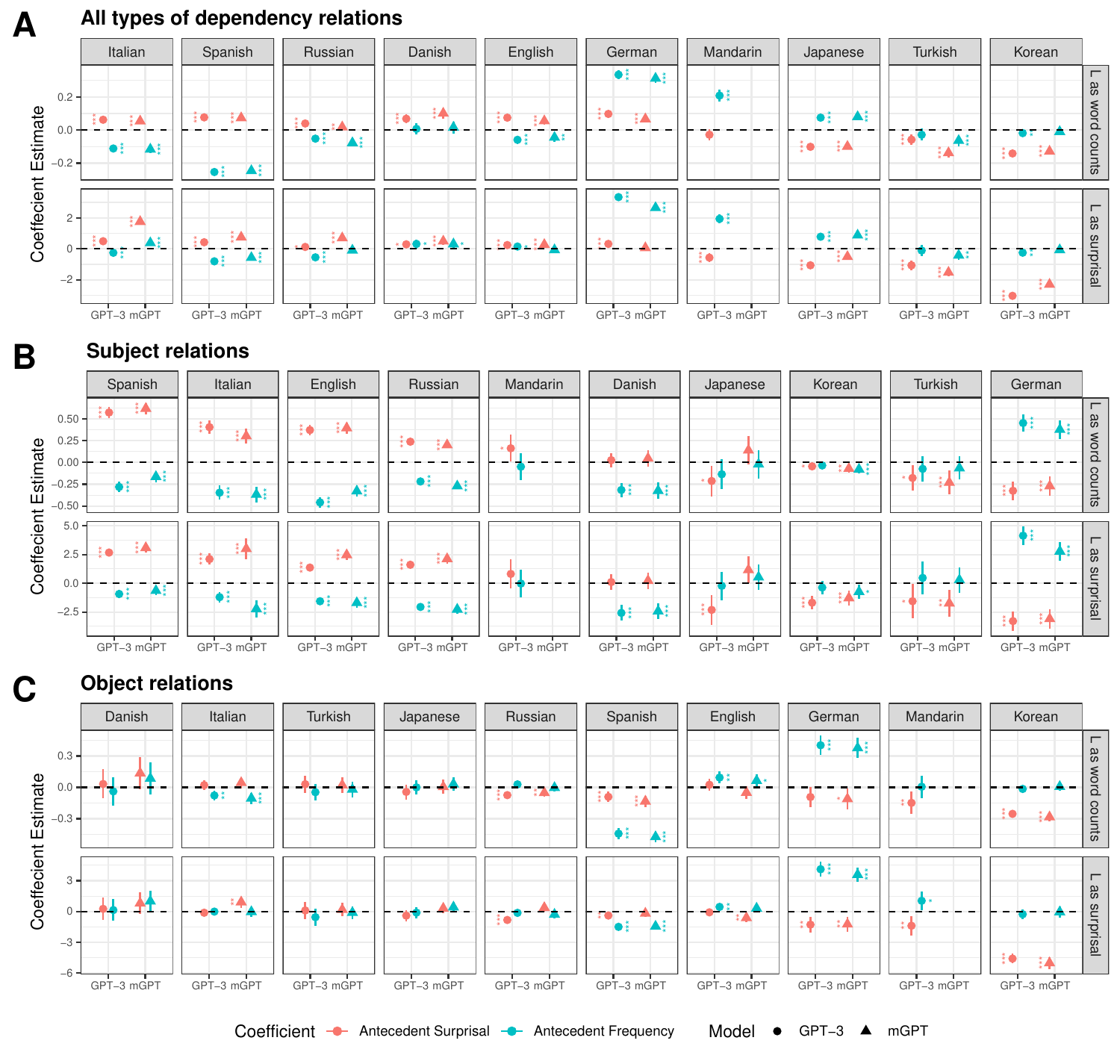}
    \caption{Study~1 coefficient estimates for the effects of \textsc{antecedent surprisal} and \textsc{antecedent frequency} on dependency length $L$ across languages, with 95\% confidence interval. Our hypothesis of strategic resource allocation predicts that more surprising antecedents are associated with longer $L$ (positive effect of \textsc{antecedent surprisal}), and that more frequent antecedents are associated with shorter $L$ (negative effect of \textsc{antecedent frequency}). Significance levels: *($p$$<$0.05), **($p$$<$0.01), ***($p$$<$0.001)}
    \label{fig:Study-1-stats-coef}
\end{figure}

\begin{figure}[]
    \centering
    \includegraphics[width=0.95\linewidth]{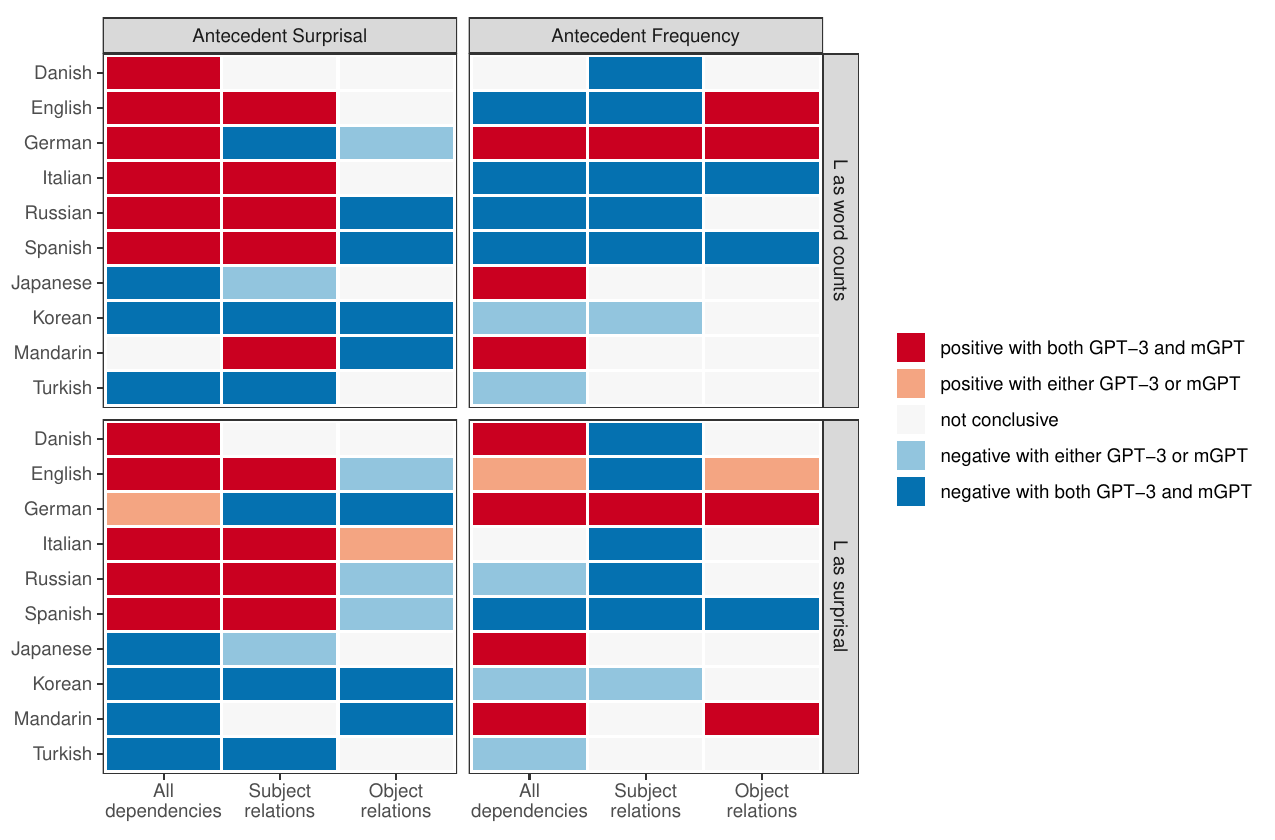}
    \caption{Study~1 summary of statistical result for the effects of \textsc{antecedent surprisal} and \textsc{antecedent frequency} on dependency length $L$. Significant ($p$$<$0.05) positive effects are highlighted in red; significant negative effects are highlighted in blue. Effects are considered not conclusive if insignificant with both language models, or if GPT-3 and mGPT show conflicting result where the effect is significant in opposite directions. According to our hypothesis, \textsc{antecedent surprisal} is expected to have a positive effect on $L$, whereas \textsc{antecedent frequency} is expected to show negative effect.}
    \label{fig:Study-1-stats-summary}
\end{figure}

The result of the raw data in its original scale is presented in Figure~\ref{fig:Study-1-DL}, which shows dependency length $L$ as a function of \textsc{antecedent surprisal}. The statistical result of the regression models is presented in Figure~\ref{fig:Study-1-stats-coef} for the effects of \textsc{antecedent surprisal} and \textsc{antecedent frequency}. A high-level summary of the statistical evidence across languages is shown in Figure~\ref{fig:Study-1-stats-summary}. Since we used the surprisal measure from GPT-3 and mGPT, we describe an effect as robust and independent of model parameterization if it is significant in the same direction both in the analysis with GPT-3 and in the one with mGPT (i.e., both positive or both negative, highlighted in dark red and dark blue in Figure~\ref{fig:Study-1-stats-summary}).\footnote{Since Mandarin is not available for mGPT, a critical effect is highlighted in dark red or blue in Figure~\ref{fig:Study-1-stats-summary} even though we only have the result with GPT-3.} We describe the effect as partially confirmed and less robust if it reaches significance with only one of the language models (highlighted in light red and blue in Figure~\ref{fig:Study-1-stats-summary}). We describe the effect as inconclusive if it is not significant with any model, or if GPT-3 and mGPT show significantly conflicting result (i.e., significantlt positive in one model but significantly negative in the other).\footnote{The use of these terms (i.e., \textit{robust}, \textit{less robust}, \textit{partially confirmed}, and \textit{inconclusive}) is only for expository purpose, and does not imply any direct statistical robustness test.}

\subsubsection{All Types of Relations}

\paragraph{Antecedent Surprisal.} In the analysis of the full dataset with all types of dependency relations, we indeed found a significant positive effect of \textsc{antecedent surprisal} for six out of ten languages, whereby more surprising antecedents are associated with longer dependency length. Specifically, for both measures of dependency length $L$, there is a positive effect in Danish, English, German, Italian, Russian, and Spanish. However, for Japanese, Korean, Mandarin and Turkish, contrary to our prediction, there is a negative \textsc{antecedent surprisal} effect.

\paragraph{Antecedent Frequency.} The result is quite mixed for the effect of \textsc{antecedent frequency}. When $L$ is measured as intervening word counts, there is a negative effect of \textsc{antecedent frequency} on $L$ in six languages (robust for English, Italian, Russian, and Spanish, while partially supported in Korean and Turkish). The effect, however, is unexpectedly positive in German, Japanese, and Mandarin, and is inconclusive in Danish. When $L$ is measured as surprisal, the \textsc{antecedent frequency} effect unexpectedly turns out to be positive for more languages, namely Danish, English, German, Japanese and Mandarin. There is still a negative effect for Russian, Spanish, Korean and Turkish.

\subsubsection{Subject Relations}

\paragraph{Antecedent Surprisal.} For subject relations, when $L$ is measured as word counts, there is a positive effect of \textsc{antecedent surprisal} on $L$ as predicted in five languages (English, Italian, Russian, Spanish, and Mandarin), suggesting that more surprising antecedents are associated with longer $L$ in these languages. However, contrary to our prediction, the effect is negative for German, Japanese, Korean, and Turkish, while Danish shows inconclusive result. Similar pattern was observed when $L$ is measured as intervening surprisal, except for Mandarin whose effect becomes inconclusive.

\paragraph{Antecedent Frequency.} The effect of \textsc{antecedent frequency} is the same for both measures of $L$. That is, there is a negative effect of \textsc{antecedent frequency} on $L$ in six languages (Danish, English, Italian, Russian, Spanish, and Korean), suggesting that more frequent antecedents lead to shorter $L$ in these languages. The effect is unexpectedly positive in German, and is inconclusive for Japanese, Mandarin, and Turkish.

\subsubsection{Object Relations} 

\paragraph{Antecedent Surprisal.} Surprisingly, for object relations, there is no positive effect of \textsc{antecedent surprisal} on $L$ in any language when $L$ is measured as word counts. Instead, there is a negative effect in German, Russian, Spanish, Korean, and Mandarin, and the result is inconclusive for the rest of the languages. Similar pattern was observed when $L$ is measured as surprisal, except that in Italian a positive effect is partially supported, and that the originally negative effect with orthographic $L$ in Russian and Spanish becomes less robust.

\paragraph{Antecedent Frequency.} There is also a mixed picture for the effect of \textsc{antecedent frequency} in object relations. When $L$ is measured as word counts, we only found a robust \textsc{antecedent frequency} effect on $L$ in four languages, two negative (Italian and Spanish) and two positive (English and German). The result for the rest of the languages is inconclusive. When $L$ is measured as intervening surprisal, there are three languages that show an unexpectedly positive effect (English, German, and Mandarin). Only in Spanish did we observe a negative \textsc{antecedent frequency} effect. The result for the rest of the languages remains inconclusive.

\subsection{Discussion}

In this cross-linguistic corpus study, we indeed found emerging evidence for a positive effect of antecedent surprisal on dependency length $L$, with both measures of $L$ showing similar patterns. This effect still holds when we zoom into the subset that only includes subject or object relations. Overall, in many languages (especially Indo-Europeans), as predicted, this pattern indicates that more surprising antecedents are associated with longer dependency length, suggesting that the pressure to minimize dependency length is relaxed when the antecedent is of higher surprisal. Consistent with our hypothesis of strategic resource allocation, the result supports that novel and unexpected linguistic units can tolerate longer dependency length before its retrieval site, possibly because unexpected information is prioritized for working memory resources during encoding, and is more resistant to memory decay and interference.

However, there are two caveats worth noting. First, there is considerable cross-linguistic variability in our result, and the antecedent surprisal effect mostly exists within Indo-European and head-initial languages in our analysis. Second, although the analysis on the full dataset with all types of dependencies reveals a general trend for a positive antecedent surprisal effect, the result is much more consistent within subject relations. In object relations, the expected effect is reversed for most languages.

It is also worth noting that the positive \textsc{antecedent surprisal} effect on $L$ cannot be reduced to a pure frequency effect. That is, there is still a significant effect of \textsc{antecedent surprisal} even though \textsc{antecedent frequency} has been included in the regression models as a control variable. Moreover, compared to \textsc{antecedent surprisal}, the effect of \textsc{antecedent frequency} on $L$ is less consistent.

In the end, to what extent does written corpus text approximate language production? Compared to spoken language, written language typically allows ``speakers'' more time to think, reducing much of the cognitive load involved in production, and the communicative goal is more geared towards listeners' need. That being said, speaker-oriented cognitive constraints, such as memory capacity, may play a less prominent role, and the need for strategic memory allocation may be diminished in written language production. Therefore, the effect observed in the current analysis using written text can be viewed as an lower bound, and we expect the effect of strategic memory allocation to be stronger when using spoken language corpora.

\section{Study 2a: Comprehension Side (Self-Paced Reading)}

\begin{figure}[t]
    \centering
    \includegraphics[width=1\linewidth]{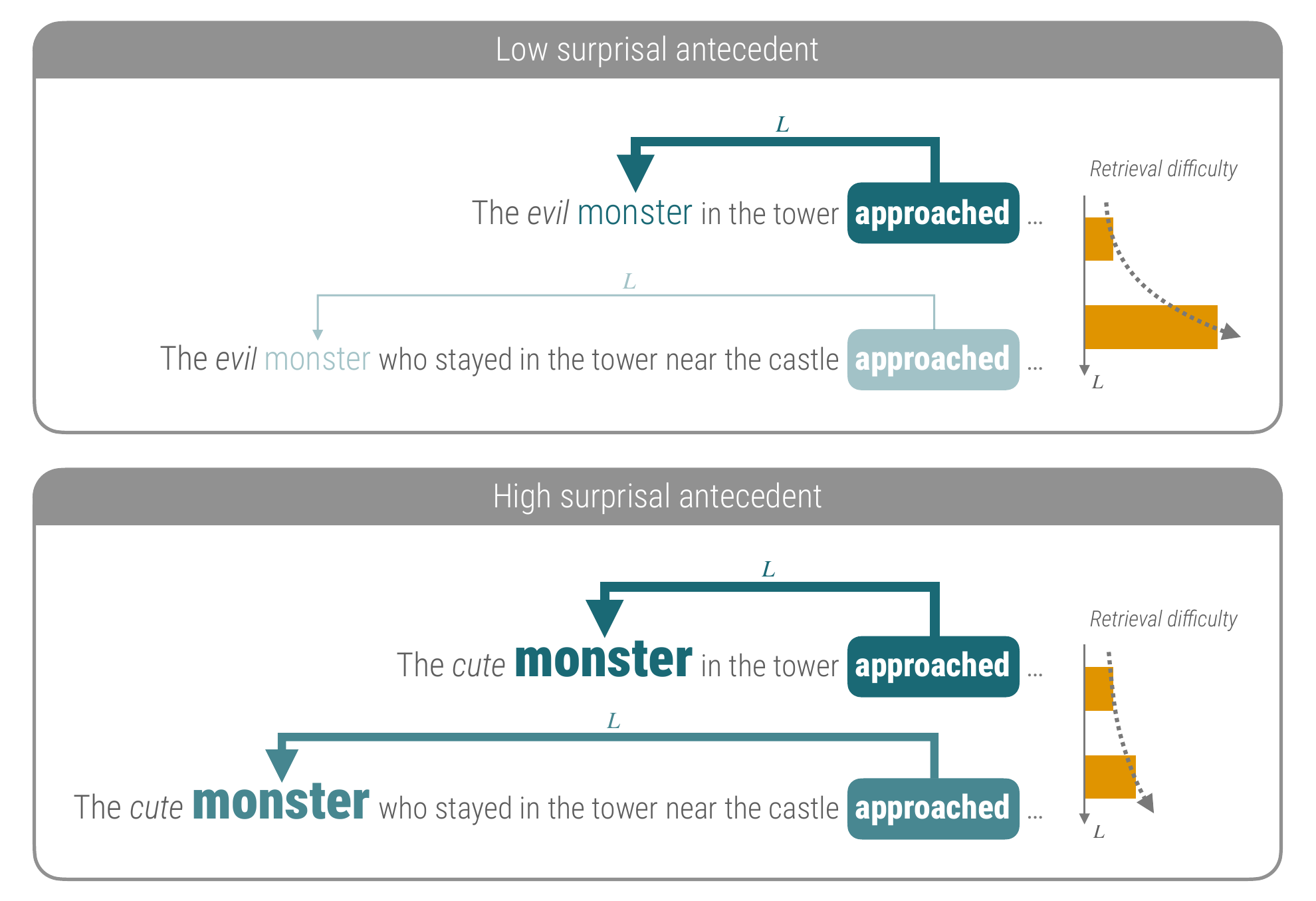}
    \caption{Empirical prediction of strategic resource allocation in comprehension. Longer dependency length leads to higher processing difficulty at the retrieval site, and this retrieval difficulty increases more slowly for high-surprisal antecedents.}
    \label{fig:study2-demo}
\end{figure}

In this second study, we investigate whether the effect of strategic resource allocation also holds from the comprehension side. In particular, we examine to what extent the dependency locality effect observed in previous comprehension studies \citep{gibson1998linguistic, gibson2000dependency, grodner2005consequences, bartek2011search} can be modulated by the surprisal of the antecedent, as illustrated in Figure~\ref{fig:study2-demo}. First, we expect a baseline dependency locality effect, where the processing difficulty at the retrieval site, manifested as reading time, is expected to increase as the dependency length gets longer. Second, according to our hypothesis, more surprising antecedents are more capable of tolerating stronger memory interference. That is, the longer dependency length does not create too much additional processing difficulty for the retrieval of surprising antecedents, resulting in a reduced locality effect.

Consider the example in Figure~\ref{fig:study2-demo}. As explained in Study~1, the more surprising ``cute monster'' in Figure~\ref{fig:study2-demo} (bottom) is more prioritized with encoding resources for its lower predictability, making it capable of tolerating longer dependency length. Therefore, when there is longer dependency distance, the processing difficulty at the retrieval site (i.e., the main verb ``approached'' in this example) should increase at slower rate for high surprisal antecedents than for low surprisal ones (Figure~\ref{fig:study2-demo}; top), since the additional intervening material induces lower level of interference for more surprising antecedents. As a result, on top of the baseline dependency length effect on retrieval difficulty, we expect a negative interaction between dependency length $L$ and \textsc{antecedent surprisal} at the retrieval site.

\subsection{Method}

\subsubsection{Data}

The data we used in Study~2a is taken from the Natural Stories Corpus (NSC) \citep{futrell2021natural}. The text of the corpus is in English, and contains 10,245 lexical words in 485 sentences, taken from 10 stories with around 1000 words each. The reading time (RT) data was collected from 181 native English speakers, using the self-paced reading task (SPR). The original corpus already excluded participants with low comprehension accuracy, as well as the reading times either faster than 100ms or slower than 3000ms. Therefore, we did not perform additional exclusion of reading time data in the current study. We generated the surprisal estimates for each word from mGPT.\footnote{GPT-3 is no longer accessible from OpenAI since January 4th, 2024. Therefore, we only used the surprisal estimates from mGPT for the analyses in Study~2a and 2b.} The NSC corpus comes with dependency annotation in UD style.

\subsubsection{Data Transformation, Exclusion, and Analysis}

As in Study~1, we analyzed three RT datasets as well here in the current study, namely the full dataset with all types of dependencies, a subset with subject relations only, and a subset with object relations only. The sample size is summarized in Table~\ref{tab:Study2-sample-size}. We ran linear mixed-effect models on the log-transformed RTs of two regions, the critical region and the spillover region. The critical region is the right codependent of each dependency, which is considered the retrieval site for the antecedent. The spillover region is the word that goes immediately after the critical region. The critical effect is the interaction between dependency length $L$ and \textsc{antecedent surprisal}, with maximal converging random intercept and random slopes by participant. For the analysis of the full dataset, we also included maximal converging random effects by dependency type. 
\begin{table}
    \centering
    \begin{tabular}{lrrrr}
        \toprule
                           &          & & \multicolumn{2}{c}{Study~2b} \\
                           \cline{4-5}
                           & Study~2a & & First-pass & Total RT \\
        \midrule
        All types          & 601,122 & & 256,567 & 313,167 \\
        Subject relations  & 57,023  & & 21,709 & 26,885 \\
        Object relations   & 48,091  & & 20,361 & 24,990 \\
        \bottomrule
    \end{tabular}
    \caption{RT data sample size in Study~2a and 2b}
    \label{tab:Study2-sample-size}
\end{table}

The control variables applied in Study~1 (namely, \textsc{sentence position}, \textsc{antecedent position}, \textsc{sentence length}, and \textsc{antecedent frequency}) are also included here in Study~2a. Besides these, we included several additional control variables that are often considered highly relevant for reading time measures. First, we included \textsc{word length}, \textsc{word surprisal} and \textsc{word frequency} of the right codependent itself. Second, to control the spillover effect often found in reading studies, we included \textsc{word surprisal} and \textsc{word frequency} of the two previous words before the right codependent. As in Study~1, frequency measures are in log scale, generated from \cite{wordfreq}. All variables are $z$-scaled. The transformation and exclusion of the dependency data follow the same procedure as in Study~1.

\subsection{Results}

\begin{figure}[t]
    \centering
    \includegraphics[width=1\linewidth]{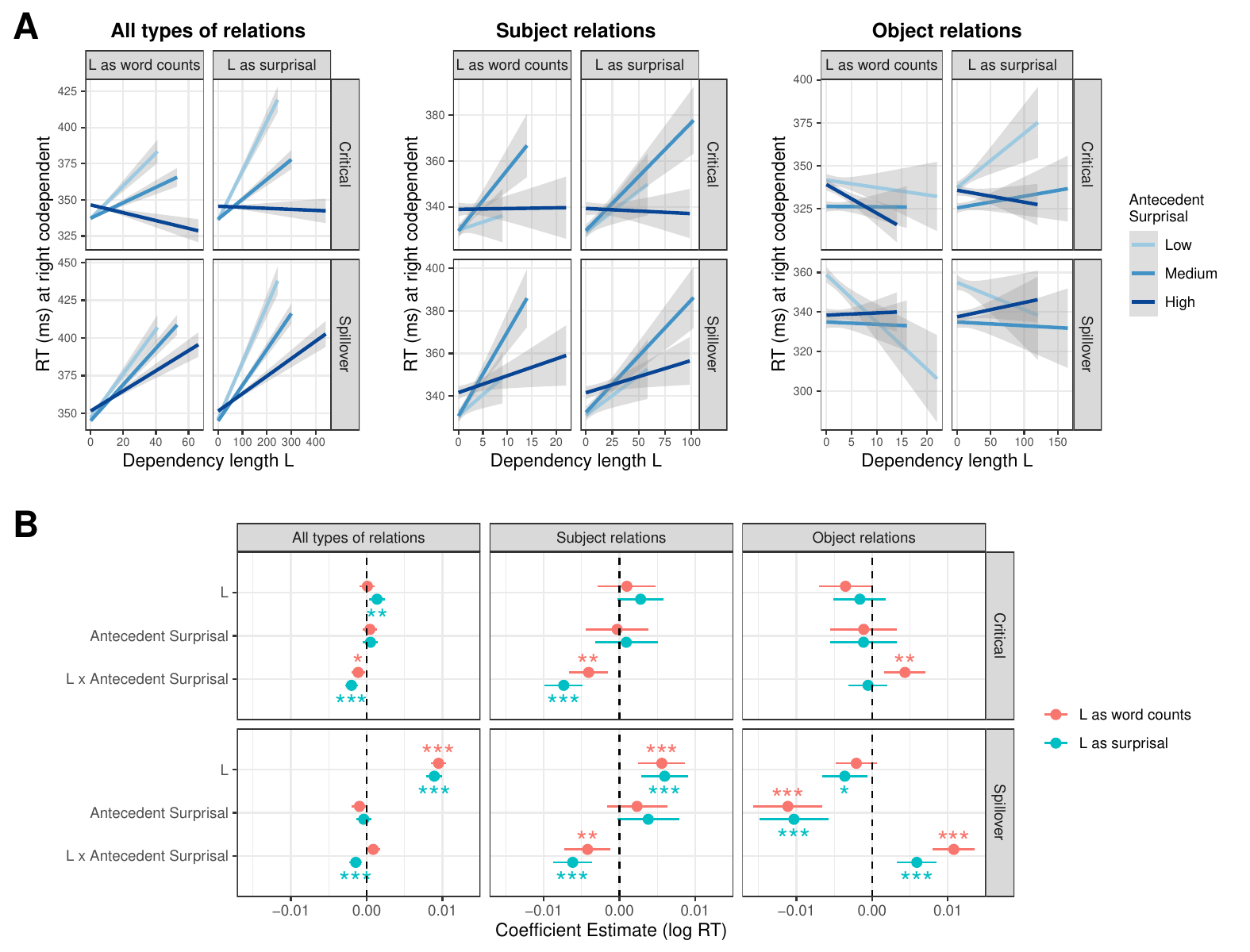}
    \caption{Study~2a reading time (RT) result. \textit{\textbf{Panel A}}: raw RTs at the right codependent and its spillover region as a function of dependency length $L$ modulated by \textsc{antecedent surprisal}, which is binned into tertiles for visualization. \textit{\textbf{Panel B}}: result of regression models with log-transformed RTs; coefficient estimates with 95\% confidence interval for the effects of $L$, \textsc{antecedent surprisal}, and the interaction between the two. Significance levels: *(p$<$0.05), **(p$<$0.01), ***(p$<$0.001)}
    \label{fig:Study2-RT-NSC}
\end{figure}

Figure~\ref{fig:Study2-RT-NSC}A shows the interaction effect between dependency length $L$ and \textsc{antecedent surprisal} on raw reading times. The result of statistical models for the critical effects is summarized in Figure~\ref{fig:Study2-RT-NSC}B.

\subsubsection{Critical Region}

\paragraph{All types of relations.} In the analysis of the full dataset with all types of dependency relations, we found a baseline locality effect where dependency length $L$ has a positive main effect on RTs at the retrieval site (i.e., the right codependent), suggesting that longer distance between codependents makes it more difficult to retrieve the antecedent at the right codependent. This main effect of $L$ is only significant when $L$ is measured as intervening surprisal. However, there is no significant main effect of \textsc{antecedent surprisal}. Importantly, we found a negative $L$ $\times$ \textsc{antecedent surprisal} two-way interaction for both $L$ measures. Consistent with our prediction, this negative interaction suggests that the locality effect on the RT of right codependents is reduced when the antecedent is more surprising.

\paragraph{Subject relations.} For subject relation, although the main effect of $L$ is numerically positive in the critical region, it is not significant with any measures of $L$. There is no \textsc{antecedent surprisal} main effect, either. However, there is indeed a significant negative $L$ $\times$ \textsc{antecedent surprisal} two-way interaction for both $L$ measures, suggesting a reduced locality effect for high surprisal antecedents.

\paragraph{Object relations.} Again, for object relations, there is no main effect of $L$ or \textsc{antecedent surprisal} for any measures of $L$ in the critical region. It is also worth noting that the $L$ main effect is numerically negative, pointing to an anti-locality effect, although this effect is not statistically significant. Surprisingly, there is a positive $L$ $\times$ \textsc{antecedent surprisal} two-way interaction, although this effect is only significant when $L$ is measured as word counts.

\subsubsection{Spillover Region}

\paragraph{All types of relations.} In the spillover region, we first found a baseline locality effect, where dependency length $L$ leads to longer RT. This baseline locality effect holds for both measures of $L$. However, similar to the critical region, there is no evidence for an \textsc{antecedent surprisal} main effect. In the end, again similar to the critical region, we found a negative $L$ $\times$ \textsc{antecedent surprisal} two-way interaction, suggesting that the locality effect is reduced when the antecedent is more surprising. This two-way interaction, however, only holds when $L$ is measured as surprisal.

\paragraph{Subject relations.} First, unlike the critical region, a baseline locality main effect of $L$ was found for subject relations in the critical region, where longer $L$ leads to longer RT at the right codependent. This $L$ main effect is significant for both measures of dependency length $L$. Second, as in the critical region, there is no \textsc{antecedent surprisal} main effect with either measure of $L$ in the spillover region. In the end, we found a critical negative $L$ $\times$ \textsc{antecedent surprisal} two-way interaction with both $L$ measures. As predicted, this negative interaction is indicative of a reduced locality effect for more surprising antecedents.

\paragraph{Object relations.} The result of object relations has a relatively complex pattern. First, there is an unexpected negative, not positive, main effect of $L$ on the RT at retrieval site. This negative main effect reaches significance when $L$ is measured as surprisal, and still numerically holds when $L$ is measured as word counts. Instead of a baseline locality effect, this negative $L$ effect suggests an \textit{anti}-locality effect, where more intervening material between the two codependents leads to faster RT at the retrieval site. Moreover, as seen in Figure~\ref{fig:Study2-RT-NSC}A (right column), this \textit{anti}-locality effect is mostly driven by antecedents of low surprisal. Second, there is a negative \textsc{antecedent surprisal} main effect for both $L$ measures, whereby more surprising antecedents induce faster RT at the retrieval site. In the end, unlike the negative $L$ $\times$ \textsc{antecedent surprisal} interaction observed in the previous two analyses, object relations exhibit a positive $L$ $\times$ \textsc{antecedent surprisal} interaction, which is significant for both $L$ measures. However, since the main effect of $L$ is negative in the first place, instead of an enhanced locality effect, this positive interaction actually suggests a reduced \textit{anti}-locality effect for more surprising antecedents. It is not yet entirely clear to us why there is an \textit{anti}-locality effect in the first place exclusively for object relations, but the result pattern in object relations seems to point to a potential trade-off between the direction of the $L$ main effect and the direction of the $L$ $\times$ \textsc{antecedent surprisal} interaction, which we will discuss below.

\subsection{Discussion}\label{sec:Study2a-discussion}

To sum up, in Study~2a, through the analysis of the RT data in the Natural Stories Corpus, we first replicated the baseline locality effect in the analysis of all types of dependency relations and subject relations, especially in the spillover region. This suggests that the non-local retrieval of the antecedent at the right codependent becomes more difficult when there is more intervening material. However, we also found an \textit{anti}-locality effect in object relations in the spillover region, which suggests that more intervening material actually facilitates the establishment of a non-local object relation. Second, only in object relations did we observe a negative main effect of \textsc{antecedent surprisal} on the RT at the right codependent, but this effect only emerges in the spillover region. This negative \textsc{antecedent surprisal} main effect suggests that more surprising antecedents are easier to retrieve in object relations, which is consistent with the findings in \cite{hofmeister2011representational}, where semantically more complex noun phrases are easier to be retrieved. However, more future work is needed to investigate why this effect only emerges in object relations. In the end, the most important finding of this Study~2a is the interaction between dependency length and antecedent surprisal, both in the critical and the spillover regions. Specifically, in the analysis of the full dataset and the one of subject relations, we found a reduced locality effect for more surprising antecedents. This reduced locality effect is consistent with the prediction of strategic resource allocation, in that more surprising antecedents are prioritized for working memory resources and are encoded with more robust representation against memory interference and decay.

It is also worth noting that there seems to be a potential trade-off between the direction of the $L$ main effect and the direction of the $L$ $\times$ \textsc{antecedent surprisal} interaction. On the one hand, when the main effect of $L$ is positive, there is a negative $L$ $\times$ \textsc{antecedent surprisal} interaction, suggesting a reduced locality effect for more surprising antecedents. On the other hand, when the main effect of $L$ is negative to start with, as in the object relations, the interaction becomes positive, indicating that the \textit{anti}-locality effect is reduced for more surprising antecedents. The reduced locality effect is straightforward, as predicted by our hypothesis. But why is there a reduced \textit{anti}-locality effect? In fact, the reduced \textit{anti}-locality effect for more surprising antecedents can be consistent with our strategic resource allocation as well. According to experience-based processing theories, the \textit{anti}-locality effect can be viewed as a facilitation effect on the prediction of the right codependent. That is, more intervening material may provide more information about the identity of the word at the right codependent, helping the comprehender to make better predictions \citep{levy2008expectation}, canceling out the burden created by memory interference. However, for more surprising antecedents, if their representation is more enhanced due to strategic resource allocation, it is possible that comprehenders can already rely on the their memory of the antecedent to predict the right codependent. As a result, the intervening material may no longer provide too much additional help to make predictions. This reduced facilitation from the intervening material for more surprising antecedent, therefore, may manifest itself as a reduced \textit{anti}-locality effect.

\section{Study 2b: Comprehension Side (Eye-Tracking)}

In Study~2b, we examine strategic resource allocation in dependency locality using an eye-tracking corpus. As in Study~2a, we expect to see a baseline locality effect, as well as an interaction between locality and antecedent surprisal, in the sense that a reduced locality effect is associated with more surprising antecedents.

\subsection{Method}

\subsubsection{Data}

The data we used in Study~2b is taken from the English part of Dundee corpus \citep{kennedy2005parafoveal}. The corpus consists of 20 texts, with 56,212 tokens in total (around 2800 words for each text). The eye-tracking data is collected from 10 English native speakers, with each text being split into 40 fine-line screens. We analyzed two eye-tracking measures: \textit{first-pass reading time}, defined as the sum of all the fixations on a region after first entering in the region and before first leaving it either to the left or to the right; and \textit{total reading time}, defined as the sum of all the fixations on a region throughout a trial. Like NSC corpus, the Dundee corpus also comes with UD-style dependency annotation.

\subsubsection{Data Transformation, Exclusion, and Analysis}

The transformation and exclusion of dependency data follow the same procedure as in Study~1 and Study~2a. The reading time responses are excluded if shorter than 100ms or longer than 3000ms. The sample size after data exclusion is summarized in Table~\ref{tab:Study2-sample-size}. We ran linear mixed-effects models on first-pass durations and total reading times. Both reading time measures are log-transformed. As in Study~2a, the critical effect is the interaction between dependency length $L$ and \textsc{antecedent surprisal}, with the same random structure and control variables as in Study~2a.

\subsection{Results}

\begin{figure}[t]
    \centering
    \includegraphics[width=1\linewidth]{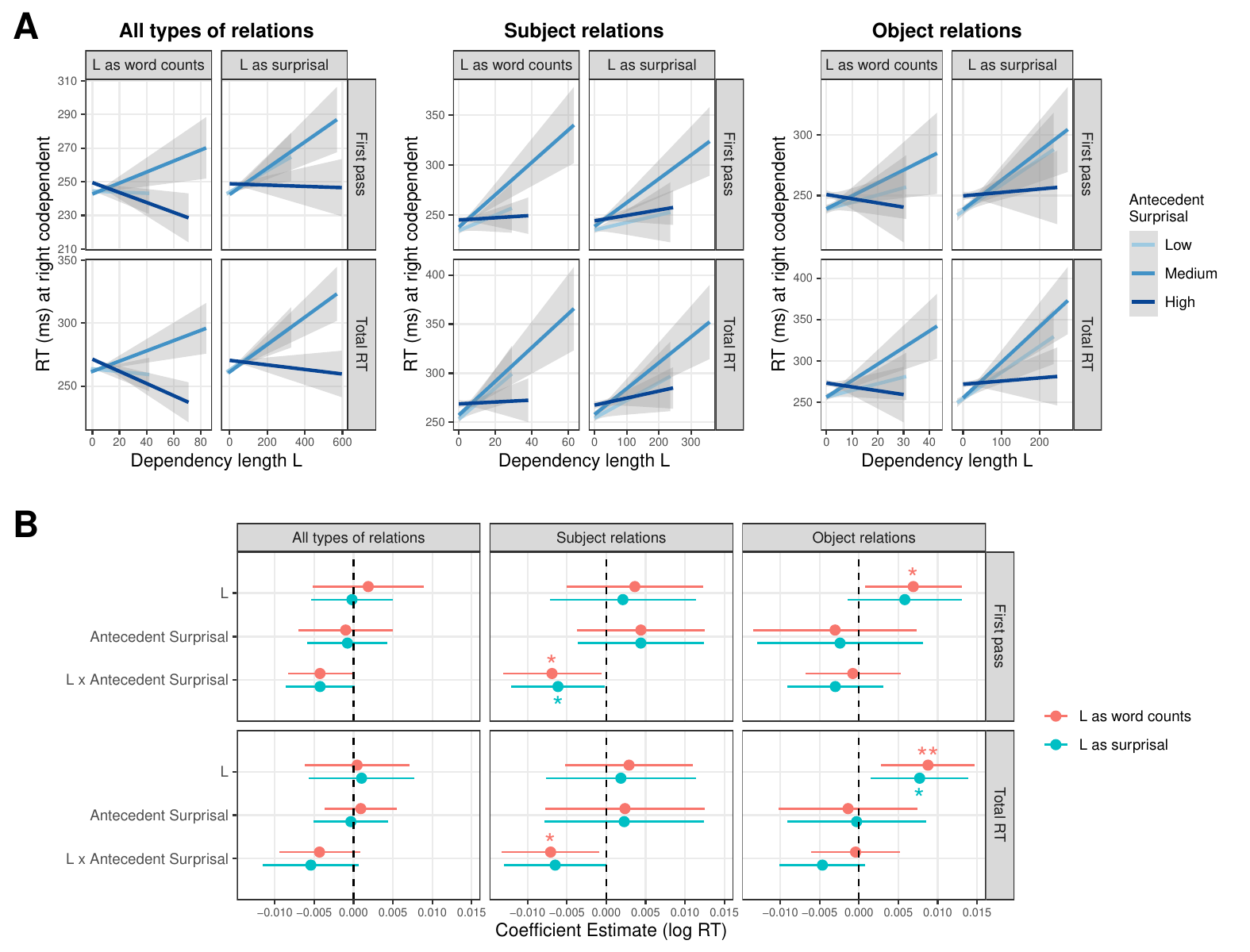}
    \caption{Study~2b reading time (RT) result for first-pass duration and total RT. \textit{\textbf{Panel A}}: raw RTs at the right codependent as a function of dependency length $L$ modulated by \textsc{antecedent surprisal}, which is binned into tertiles for visualization. \textit{\textbf{Panel B}}: result of regression models with log-transformed RTs; coefficient estimates with 95\% confidence interval for the effects of $L$, \textsc{antecedent surprisal}, and the interaction between the two. Significance levels: *(p$<$0.05), **(p$<$0.01), ***(p$<$0.001)}
    \label{fig:Study2-RT-Dundee}
\end{figure}

Figure~\ref{fig:Study2-RT-Dundee}A shows the interaction effect between dependency length $L$ and \textsc{antecedent surprisal} on raw reading times, including both the first-pass duration and the total RT. The result of statistical models for the critical effects is summarized in Figure~\ref{fig:Study2-RT-Dundee}B.\footnote{No critical effects were found in the spillover region, so we only report the result of the critical region in this Study~2b.}

\paragraph{All types of relations.} As shown in Figure~\ref{fig:Study2-RT-Dundee}B (left column), no evidence was found for the main effect of dependency length $L$ with any measure of RT and $L$, suggesting the lack of the baseline locality effect. No significant main effect of \textsc{antecedent surprisal} was found, either. For the critical $L$ $\times$ \textsc{antecedent surprisal} interaction, although the effect is numerically negative on first-pass RT with both measures of $L$, it is only marginally significant. No evidence for the interaction effect was found on total RT.

\paragraph{Subject relations.} Still, as shown in Figure~\ref{fig:Study2-RT-Dundee}B (middle column), we did not find any evidence for either an $L$ or an \textsc{antecedent surprisal} main effect with any measure of RT and $L$. However, there is indeed a negative $L$ $\times$ \textsc{antecedent surprisal} two-way interaction, suggesting that the locality effect at the retrieval site, although not statistically significant on average, is reduced for more surprising antecedents. This interaction effect reliably holds for first-pass RT with both $L$ measures, as well as for total RT with $L$ measured as word counts. It is, however, only marginally significant for total RT with $L$ as intervening surprisal.

\paragraph{Object relations.} As shown in Figure~\ref{fig:Study2-RT-Dundee}B (right column), we found in object relations a baseline locality effect manifested as a positive $L$ main effect on RTs at the retrieval site, which holds for first-pass RT with $L$ as word counts and for total RT with both $L$ measures. There is no evidence for an \textsc{antecedent surprisal} main effect. In terms of the critical $L$ $\times$ \textsc{antecedent surprisal} interaction, we only found a marginally significant negative interaction for total RT with $L$ measured as surprisal, and the effect is non-significant elsewhere.

\subsection{Discussion}

The result of the two main effects in the current Study~2b shows a very different pattern compared to Study~2a. First, unlike Study~2a, in the current Study~2b only in object relations did we find a baseline locality effect, whereas the locality effect is not observed in the analysis of the full dataset or in the one of subject relations. That is, longer dependency length does not lead to higher processing difficulty at the retrieval site for subject relations. This lack of the baseline locality effect aligns with the observation in \cite{demberg2008data}, where the locality effect in Dundee corpus is overall small and unreliable for verbs, which in our case is the retrieval site of subject relations. Second, there is no main effect of \textsc{antecedent surprisal} across the board in this Study~2b. 

Although the baseline locality effect is relatively unreliable, we still observed evidence for a negative $L$ $\times$ \textsc{antecedent surprisal} interaction effect in this Study~2b, especially in subject relations. As shown in Figure~\ref{fig:Study2-RT-Dundee}A (middle column), compared to antecedents with low-to-mid surprisal levels, those with high surprisal exhibit weaker locality effect. Similar to the interaction effect observed in Study~2a, the current result shows that more surprising antecedents is less susceptible to the effect of memory interference induced by intervening material, possibly due to their enhanced representation. Supplementing the self-paced reading data in Study~2a, the result of Study~2b thus lends support to our hypothesis of strategic resource allocation with data from eye-tracking paradigm.

\section{General Discussion}

In three corpus studies, we examined strategic resource allocation (SRA) through the lens of dependency locality both in production and in comprehension. Study~1 explored this hypothesis in production by analyzing UD corpora of 10 languages. Our result reveals that more surprising antecedents can tolerate more intervening material before they need to be retrieved at the other side of the dependency, resulting in a positive correlation between antecedent surprisal and dependency length. However, it is worth noting that this reduced locality effect mostly exists within Indo-European and head-initial languages, and is more consistent for dependencies of subject relations than for object relations. 

In Study~2, we shifted gears and focused on comprehension, analyzing two English reading-time corpora: one based on self-paced reading paradigm (Study~2a) and the other based on eye-tracking paradigm (Study~2b). The SPR data reveals a baseline locality effect, where longer dependency lengths lead to increased reading times at the retrieval site. Importantly, we found a $L$ $\times$ \textsc{antecedent surprisal} interaction, where the baseline locality effect is reduced for more surprising antecedents. Moreover, we again observed a subject-object asymmetry, such that the critical effect consistently holds only in subject relations. The eye-tracking data shows a more nuanced pattern: although the baseline locality effect was observed only in object relations, we indeed found an $L$ $\times$ \textsc{antecedent surprisal} interaction in subject relations.

Overall, despite the caveats mentioned above, our result shows emerging evidence that a reduced locality effect emerges for more surprising antecedents in the processing of non-local dependencies, suggesting that more surprising antecedents are less susceptible to the interference from intervening material. This finding aligns with the notion of strategic resource allocation that we proposed, which holds that unexpected information is prioritized for memory resources and is encoded with enhanced memory representation.

\subsection{Processing difficulty as encoding difficulty: Reinterpreting the effect of surprisal and entropy}\label{sec:encoding-diff}

\begin{figure}[htp]
    \centering
    \includegraphics[width=0.5\linewidth]{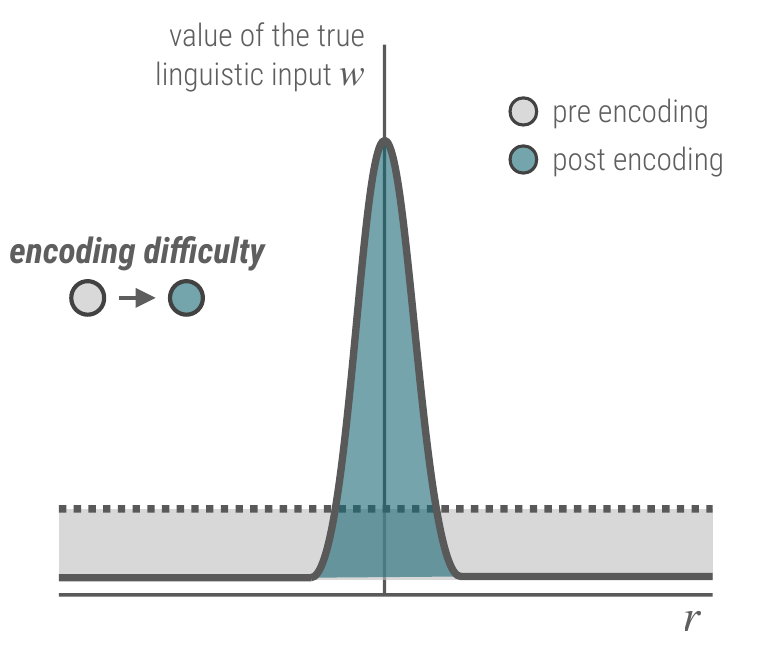}
    \caption{Processing difficulty as memory encoding difficulty}
    \label{fig:encode-diff}
\end{figure}

In this section, we first reinterpret the processing difficulty of a word as its memory encoding difficulty. As shown in Figure~\ref{fig:encode-diff}, the encoding process in Figure~\ref{fig:SRA-demo} can be considered a transformation from a flat uniform distribution over all possible words to one that is concentrated around the received input. The processing difficulty of a word at the encoding stage, therefore, can be considered the distance between the pre-encoding and the post-encoding distribution. As a result, less uncertain, or more precise encoding distribution is more distant away from the uniform pre-encoding distribution, leading to higher processing difficulty. With this encoding view in mind, let us consider two factors that have been previously argued to influence the processing difficulty of a word, namely surprisal and entropy.

For the surprisal effect, as shown in our main proposal of SRA, more surprising input should be encoded with higher precision, an efficiency strategy that we have argued to minimize the retrieval error at a later time point. As a result, SRA naturally predicts that the more precise encoding for more surprising inputs should lead to higher encoding or processing difficulty, which is consistent with the widely observed surprisal effect in the literature. Importantly, this memory encoding view provides a resource-rational account for the surprisal effect, reinterpreting it as a strategic solution to the efficiency problem of memory. We will discuss this in more detail below in Section~\ref{sec:surprisal-efficient-coding}.

For the effect of entropy, SRA yields complicated predictions. As mentioned in Introduction, SRA implies that higher uncertainty of prior does not necessarily increase or decrease the precision of encoding distribution, and therefore does not necessarily increase or decrease the processing difficulty of a word. To demonstrate the reason behind it, first recall that according to SRA the encoding precision depends on how the received input can be accurately reconstructed at a later time point, which is in turn dependent on to what extent such a reconstruction can be supported by the prior. Intuitively, when the prior is highly consistent with the actual received word (e.g., low surprisal words), the more precise the prior is, the less precise the encoding of that word needs to be for better reconstruction. In this case, the uncertainty of prior should have a positive effect on encoding difficulty, in the sense that the processing of a word is facilitated if its preceding context yields a high-constraining prior. However, when the prior is not compatible with the actual input (e.g., high surprisal words), such a facilitation effect is necessarily the case any more: a highly precise but incompatible prior may actually need to be counteracted by more effort into precisely encode the actual input. In fact, this complicated effect of prior precision echos the empirical observation in many previous studies, where the effect of prediction entropy is much less reliable than the effect of surprisal on processing difficulty \citep{linzen2016uncertainty, van2021single, wilcox2023testing}.

\subsection{The role of representational uncertainty} \label{sec:role-uncertainty}

Under SRA, there is a dissociation between \textit{accuracy} and \textit{uncertainty} in memory representations. For accuracy, it is more related to the \textit{mean}, or \textit{point estimate} of an underlying distribution with respect to how far it is from the true input value. For uncertainty, it corresponds to the \textit{precision}, or \textit{variance} of the distribution, reflecting the relative competitiveness of all alternative inputs. As mentioned in Introduction, the critical prediction of SRA is about the \textit{precision} or \textit{uncertainty} in the encoded memory representation, as reflected in robustness to interference, rather than about the raw \textit{accuracy} of retrieval. As shown in Figure~\ref{fig:SRA-demo}, although a relatively high retrieval accuracy may be maintained on average for all linguistic inputs, those that receive more resources will have less uncertainty in their encoded representation.

Both the point estimate and the uncertainty are important information to understand the underlying representations of memory processes, as raised by more and more recent work in psychophysics \citep{bays2024representation}. However, compared to point estimates, the characterization of representational uncertainty is relatively underexplored in psycholinguistics research, both theoretically and empirically. For example, studies under the framework of cue-based retrieval often concern what representation has actually been retrieved, based on which an empirical prediction is derived. Similarly, studies under the noisy-channel framework often focus more on interpreting the point estimate of posterior distribution, rather than how the probability mass is distributed over the hypothesis space.

One of the major challenges to understand the role of representational uncertainty is possibly the lack of a straightforward linking hypothesis. In most psycholinguistics studies, the interpretation of \textit{online} dependent measures such as reading times is based on the point estimate of its mean, which is naturally linked to the point estimate of mental representations  \citep[cf.][]{huang2023infrequent}. But for representational uncertainty, how this psychological construct is linked to any online behavioral measure remains unclear. Even though there might be some ways to probe the degree of representational uncertainty through certain \textit{offline} measures (e.g., tasks that directly probe the errors in the interpretation of a sentence), it is still challenging to lay out the hypothesis space of alternative representations in a fine-grained manner.

In the current study, an important assumption we made is that less uncertainty leads to more robust representation against interference. That is, if the processesor is more uncertain about the representation of a linguistic input, its encoding distribution may be more likely to be influenced and distorted by other information it holds in memory. Surely, this is a debatable assumption, and the specific mechanism of how the representation is distorted by other inputs needs to be further elaborated in future work.

\subsection{The role of context in working memory efficiency}

Earlier in Introduction, we have argued that the efficiency of working memory allocation depends on how likely a linguistic unit can be reconstructed based on the statistical structure of linguistic input. But an open question is: what kinds of statistics are being used? More specifically, to what extent is working memory efficiency guided by context-sensitive statistics?

The role of contextual information has been under debate across multiple domains of linguistic efficiency, especially in the context of signal reduction \citep{jaeger2017signal}. One of the earliest evidence supporting the communicative efficiency pressure for linguistic structure is the well-known Zipf's law, which observes that more frequent words tend to have short forms \citep{zipf1949human}. Similar reduction effect of linguistic forms as a function of usage frequency has also been found in the historical change of linguistic representations \citep[e.g.,][]{bybee1994evolution, bybee2006usage, pierrehumbert2008exemplar, cohen2015informativity}. While most of these theories focus on frequency, which can be considered unigram probabilities independently generated from a stationary distribution, recent studies have begun investigating the role of context-specific probabilities in linguistic efficiency. The findings in this area are mixed. For example, building on Zipf's observation, \cite{piantadosi2011word} find that a significant amount of word-length variability is explained by contextual predictability in addition to the frequency effect. In contrast, \cite{pimentel2023revisiting} argue that word length is better predicted by frequency. Beyond the structure of lexicons, in online processing, contextual predictability also shapes the reduction of referring expressions \citep[e.g.,][]{mahowald2013info, tily2009refer, xu2021there} as well as syntactic structure \citep[e.g.,][]{jaeger2010redundancy, jaeger2006speakers}.

In the current study, that the observed antecedent surprisal effect cannot be reduced to a pure frequency effect suggests that the statistics relevant to working memory efficiency go beyond unigram frequencies, and the strategic allocation of working memory resources is based on more fine-grained context-specific probabilities. As noted by \cite{jaeger2017signal}, frequency can be understood as an averaged effect of contextual predictability. This is probably one of the reasons why linguistic theories focusing on the representational aspects of language often emphasize frequency, as it reflects the abstract, global properties of a language accumulated through long-term experience. That being said, the optimization of working memory efficiency not only relies on those ready-to-retrieve statistics already stored in long-term memory, but also incorporates statistics computed online, dynamically drawing upon rich contextual information in a rapid and adaptive manner \citep[for a different view, see][]{opedal2024role}. 

Moreover, the role of context in SRA has its implications for the asymmetry between proactive and retroactive interference. Proactive interference refers to the configuration where distractor information precedes the retrieval target, whereas for retroactive interference the distractor is located between the retrieval target and the retrieval site. Previous studies have observed that proactive interference has weaker effect than the retroactive one \citep{van2011cue}, an empirical pattern that supports time-based decay of memory activation \citep{portrat2008time, barrouillet2004time, page1998primacy, lewis2005activation}. Under SRA, this asymmetry can be potentially explained from a resource-rational perspective. As shown in Figure~\ref{fig:SRA-demo}, at the encoding stage of a linguistic unit, the processor already has access to the information in its preceding context, but not yet to the upcoming information in the right context. Therefore, during memory encoding, the strategic allocation is possibly based only on the preceding context. In other words, the encoded representation of an input is optimized for information it has already received in the preceding context, but not necessarily for what has not yet been received. If this is the case, SRA naturally explains why the distractor that goes before the target unit has less impact on its representation than the retroactive distractor, without necessarily resorting to a separate time-based decay mechanism.

\subsection{Relationship with lossy-context surprisal}\label{sec:lossy-context}

The theoretical framework of SRA we proposed shares similar theoretical and empirical implications with lossy-context surprisal theory and its variants \citep{futrell2020lossy, hahn2022resource}. Focusing on the prediction mechanism, lossy-context surprisal holds that next-word predictions are based on lossy and faulty memory representations, rather than the veridical form of the past linguistic input. The theory explicitly includes a memory distortion process, where certain elements in an utterance are erased to form a lossy representation, subject to certain probabilistic erasure distributions.

In many aspects, our proposal and lossy-context surprisal form two sides of the same coin. For our SRA, we seek to understand and explain the working memory mechanism in sentence processing, and predictive processing is a component incorporated into the mechanism we proposed to better explain memory. For lossy-context surprisal, in contrast, the theory aims to understand the prediction mechanism in sentence processing, with a memory component included to better explain prediction. Despite these different goals, we both speak for an interaction between memory and prediction, as both of them should jointly support human linguistic behaviors as a cognitive task. That means, our proposal and lossy-context surprisal are not mutually exclusive, theoretically or empirically, and we simply focus on different perspectives of a similar cognitive task.

In fact, our theoretical framework of SRA is on some level mutually translatable with lossy-context surprisal, so we do not see them necessarily as conflicting theories. On the one hand, in the language of lossy-context surprisal, the memory distortion process where certain linguistic units are erased is potentially where our strategic resource allocation could fit in, such that more surprising units given the context are less likely to be erased when predicting future units. On the other hand, a fundamental question at the core of our resource-rational analysis of working memory mechanism is: if memory capacity is limited, how to minimize the cost of memory error by strategically prioritizing more important linguistic units? However, what counts as important? Or, in other words, what is the objective function based on which the cost of memory error is defined? In the original lossy-context surprisal theory \citep{futrell2020lossy}, this objective function is not explicitly specified. In \cite{hahn2022resource}, the model takes one step further, and this objective function is to minimize the downstream next-word prediction task. In our proposal, the cost is defined by how likely a unit can be reconstructed later given the context. These two different objective functions of cost are not mutually exclusive, and the accuracy of next-word prediction may be compromised if a lost unit in memory is not reconstructable. Therefore, although our proposal of SRA has a different theoretical focus from the model of \cite{hahn2022resource}, our empirical predictions actually share some overlap, and we both predict on some level that the representation of more surprising units (or, less frequent units) should be enhanced and more robust.

\subsection{Hierarchical encoding and compression}

SRA, arising as an efficiency principle from the functional pressures of working memory, is situated more at the computational level in \citet{Marr:1982:VCI:1095712}'s three-level representation. A natural question to ask then is: what is the potential mechanism to implement this efficiency principle at the algorithmic level? In other words, when more resources are allocated, what makes the representation less uncertain and more robust?

One possible mechanism is hierarchical compression, which postulates that information can be stored in memory with a multi-level hierarchy of abstraction \citep{brady2009compression, bates2020efficient, brady2024noisy, craik1972levels, christiansen2016now}. In visual working memory, higher levels are more compressed, having a more categorical nature; lower levels, in contrast, are encoded with more quantitative perceptual detail. A similar hierarchy also exists in sentence processing. Sequential linguistic input can be continuously encoded and recoded into compressed forms, which in turn are further compressed into more abstract forms when new input comes in \citep{christiansen2016now}. This incremental compression procedure gives rise to a multi-level hierarchical structure of memory representation, such that higher levels of abstraction are encoded as a gist of message without specifying elaborated syntactic and semantic features \citep{bradshaw1982elaborative}. Intuitively, more memory resources should yield more detailed encoding. Indeed, this has been recently demonstrated by some memory encoding models grounded in the rate-distortion theory, where a quantitative-categorical spectrum naturally arises simply by manipulating the memory capacity during encoding \citep{bates2020efficient, jakob2023rate}.

Predictable information, if less prioritized, should be encoded in a more compressed and abstract fashion. In visual working memory, this is indeed evidenced by the fact that more memory objects can be stored in visual working memory tasks when their perceptual features are statistically correlated \citep{brady2009compression, bates2019adaptive}. Moreover, information with stronger prior knowledge is also more susceptible to the categorical bias in perception, where the perceived input is biased towards the categorical mean \citep{bates2020efficient}. Similarly, in the domain of language, more frequent linguistic sequence is more likely to be holistically stored in memory \citep[e.g.,][]{bybee2006usage, goldberg2003constructions, hawkins2004efficiency, traugott2013constructionalization}. Importantly, hierarchical compression based on statistical regularities provides a mechanism that strings together effects across different linguistic representational levels, forming a spectrum of compression. At one end, there is the locality effect where words that are more mutually predictable tend to stay closer to each other in linear order \citep[e.g.,][]{futrell2019syntactic, futrell2019information}. At the other end, the same pressure of compression governs the fusion of morphemes \citep[e.g.,][]{hahn2021modeling, rathi2021information}, and makes mutually predictable units more likely to go through processes such as affixation and phonological reduction \citep[e.g.,][]{bybee1994evolution, bybee2006usage, gahl2024time}.

\subsection{Parallelism between production and comprehension}

The effect of SRA holds both for production and for comprehension in the current study, pointing to a parallelism between these two modalities. For comprehension, a reasonable question is: to what extent the observed effect of strategic resource allocation is experience-based \citep{macdonald2002reassessing}, given that the same effect is also seen in production data? In other words, it is possible that comprehenders prioritize unexpected linguistic units and encode them with enhanced representation because unexpected units are more likely to associate with stronger memory interference in the production data they receive. 

A similar question can be asked for production as well: to what extent is the effect observed in production the result of audience design \citep{clark1982audience, ferreira2019mechanistic, lockridge2002addressees}, given that comprehenders can encode unexpected units with prioritized memory resources? One listener-oriented production theory compatible with our finding is the Uniform Information Density (UID) theory  \citep{jaeger2006speakers, meister-etal-2021-revisiting, clark2023cross}. According to UID, surprising antecedents may be followed by longer dependency length so that there is a smoother transition to the other side of the dependency, which is relatively predictable since it shares high mutual information with the antecedent \citep{futrell2019syntactic}. 

Here we do not attempt to adjudicate between the two questions above, nor do we view them in conflict with our proposal. In our opinion, SRA provides a potential explanation for the mechanistic underpinnings of these higher-level processes.

Despite the parallelism, comprehension and production still differ in some critical aspects, exerting modality-specific constraints on SRA due to their idiosyncratic processing nature. For example, production is, in general, more cognitively demanding, in need of higher memory capacity, executive control, and action planning than comprehension \citep{macdonald2013language, nozari2017monitoring, hickok2012computational, koranda2020language}. This additional cognitive demand may exert more pressure to efficiently use the limited memory resources in production than in comprehension, possibly resulting in a stronger effect of SRA. Future work is needed to investigate this possibility.

\subsection{Implications for linguistic typology}

Dependency length minimization as a functional universal has been argued to shape the syntactic structure of human language, due to the pressure to efficiently use the limited working memory resources. In the current study, we go beyond the general constraint of limited memory capacity, and further argue that memory resources should be strategically allocated to prioritize novel and unexpected information, a memory efficiency principle that naturally arises from two assumptions about working memory. Our results indicate that this strategic resource allocation indeed serves as a functional constraint to shape syntactic structures, in the sense that the pressure to minimize dependency length can actually be relaxed when the antecedent of a syntactic dependency is of higher surprisal. Our finding further substantiates the functionalist view as a promising approach to provide explanatory accounts for linguistic universals \citep{gibson2019efficiency}. It also highlights the importance of having increasingly sophisticated characterization of functional constraints, in order to see how far we can go with this functionalist view on the structure of human language.

Despite this goal of having SRA as a universal efficiency principle to explain language structure, an important question is: how universal is SRA cross-linguistically, and how does it interact with other grammatical constraints and language-specific phrase structures? 

First of all, one consistent pattern we have observed is the asymmetry between subject and object relations. Specifically, the effect of SRA is generally less reliable for object relations than for subject relations. One possible explanation is that object relations are subject to stronger grammatical constraints, with greater pressure to position the head and its dependent closer to each other. In support of this possibility, \cite{Keenan1977noun} identify an Accessibility Hierarchy as a linguistic universal, where noun phrases in the subject position are more readily relativized into relative clauses than those in the object position. Such constraints may bind object noun phrases and verbs more tightly, reducing the influence of SRA. Moreover, in many languages, grammatical agreement is common in subject--verb dependencies but absent in object-verb dependencies. Therefore, establishing subject relations may require more active grammatical computation, resulting in increased memory demand in processing and a stronger need for more strategic and efficient use of working memory resources.

Another notable pattern in Study~1 is that, compared to head-initial languages, most of the head-final languages in our analysis do not exhibit a reliable antecedent surprisal effect on dependency length. We speculate that this may be related to the tendency for argument dropping in head-final languages with SOV word order. From the perspective of dependency locality, SOV word order is associated with longer dependency lengths compared to SVO, which should theoretically impose higher memory costs and make it a less efficient structure. Despite this inefficiency, typologically, SOV is a word order commonly attested \citep{hammarstrom2016linguistic}. As an explanation for this paradox, some studies find that arguments in SOV languages are often dropped, reducing the overall dependency length in actual language use \citep{levshina2025paradox, ueno2009does}. It is possible that by allowing speakers to drop arguments, the efficiency of SOV structure may already be significantly improved, obviating the need for SRA as another efficiency strategy.

It is also worth mentioning a few other relevant language-specific factors. For Mandarin Chinese, although it predominantly follows SVO word order, the relative clause goes before the nouns, increasing the flexibility of object relations in terms of their dependency length. For Korean and Japanese, the subject noun phrase may be delayed when it is surprising and unexpected, shifting the word order from SOV to OSV, and therefore decreasing the dependency length of subject relations for surprising antecedents.

\subsection{Surprisal effect as efficient coding: An adaptionist view}\label{sec:surprisal-efficient-coding}

One way to interpret our finding is that the enhanced robustness of memory representation arises as a by-product of the effort involved in processing surprising information. However, this raises an even more fundamental question: why does the surprisal effect occur in the first place? In other words, why is there a widely observed positive relationship between surprisal and processing effort?

Here is one way to think about the basic surprisal effect from an information-theoretic perspective. More surprising linguistic units correspond to longer code length. For example, consider a low-surprisal word encoded by a sequence of 3 bits \texttt{110}, compared to a high-surprisal word encoded by 11 bits \texttt{11100001101}. An (over-)simplified mechanical interpretation of the basic surprisal effect, therefore, is that encoding longer sequence of code in memory requires more time and effort. This leads to the widely observed linear relationship between surprisal and behavioral measures, such as reading time \citep[e.g.,][]{smith2013effect, hoover2023plausibility, shain2024large, wilcox2023testing, xu2023linearity}.

However, choosing a code length based on predictability is not the only possible strategy to encode a linguistic unit. An alternative approach is to assign each unit a code of equal length regardless of statistical regularities, a scheme analogous to the uniform distribution of resource allocation discussed in Introduction. For example, the ASCII (American Standard Code for Information Interchange) system uses such a coding strategy, where every character is represented as a 7-bit sequence. Under this hypothetical scenario, processing effort would be insensitive to statistical regularities and uniformly distributed across all linguistic units.

As demonstrated in previous work on information theory, a coding scheme that treats every input as equally likely is inherently inefficient. When the statistical structure of the input is known, a coding scheme that assigns shorter code lengths to more predictable inputs can reduce the average code length, thus increasing efficiency \citep{shannon1948mathematical}. The observed linear relationship between surprisal and processing effort suggests that the brain may adopt such an efficient coding strategy. Specifically, the brain appears to assign code lengths to input units in proportion to their likelihood of occurrence based on long-term statistical regularities. This strategy is supported by the evidence that the brain is very good at inferring and approximating the statistical structure of the external environment \citep{saffran1996statistical}, and there is good reason to believe that the brain uses these inferred statistical structures to encode information more efficiently.

The connection between the surprisal effect and efficient coding suggests that the basic surprisal effect itself may be construed in evolutionary terms. Under certain efficiency pressures, the cognitive system may have evolved to adopt a memory encoding strategy that optimizes an objective function within the constraint of limited resources. The exact nature of this objective function remains an open question: it could be the minimization of distortion cost, or it could involve something else. The key point, however, is that the relationship between surprisal and strategic resource allocation can be understood at different timescales. On the one hand, at the level of processing individual sentences, strategic resource allocation may arise as a by-product of the effort required to process surprising information. On the other hand, over a longer timescale, the tendency to invest more effort into encoding less predictable information may reflect an evolved strategy that is adapted to an efficiency problem.

\subsection{Limitations}

One major limitation of our analysis is that the results heavily depend on the quality of surprisal measures generated from LLMs. In the current study, we conducted our analysis using the surprisal from two language models, namely GPT-3 and mGPT. As presented above, the two models do yield consistent pattern, alleviating the concern that our main result may be the artifact of any model-specific behavior. However, it still does not entirely rule out the issue with the accuracy of LLM surprisals, especially for low-resource languages that are under-represented in the training data of the models we used. This limitation may compromise the extensibility of our analysis to under-studied languages, which are of particular interest from a typological perspective, and is particularly relevant for the unreliable effect we observed for some non-Indo-European languages.

In the end, even though contemporary LLMs can provide state-of-the-art probabilistic measures for linguistic data, it remains questionable to what extent it reflects the predictive processing in humans. Despite the correlation between the model-generated surprisals and the behavioral or neural responses in humans, many studies actually find that there still remain some critical patterns in human empirical data that cannot be fully accounted for solely by surprisals \citep{huang2024large, van2021single}. Moreover, compared to humans, modern LLMs are far less constrained in terms of their memory capacity. This makes LLMs less likely to resemble the memory architecture in humans, or to capture the memory processes stemmed from the efficiency pressure exerted by the limited memory capacity \citep{oh2023does, timkey2023language}. However, this is not \emph{necessarily} a limitation for the current study, since there are cases where a model with superhuman memory can provide probabilistic measures that more accurately reflect the statistical properties in the linguistic data without being confounded by the memory interference in the language model itself.

\section{Conclusion}

The current study proposes Strategic Resource Allocation (SRA) as an efficiency principle for memory encoding in sentence processing, which holds that working memory resources are strategically and dynamically allocated to prioritize novel and unexpected information. Theoretically, we argue that SRA is an efficient solution to the computational problem faced by working memory, that is, to maximize the retrieval accuracy of past inputs under the constraint of limited memory resources. Empirically, this principle predicts that the memory representation of more surprising linguistic units is more robust against interference and decay. We examined this prediction through naturalistic corpus data in the context of dependency locality from both the comprehension and the production side. In production, through the analysis of UD corpora in 10 languages, we indeed found that more surprising antecedents can tolerate longer dependency length, but the effect mostly exists within Indo-European and head-initial languages. This cross-linguistic variability therefore calls for a closer look into how SRA as a domain-general memory efficiency principle interacts with the language-specific phrase structure. In comprehension, through two English reading time corpora, we observed a similar reduced locality effect on retrieval difficulty for more surprising antecedents. Moreover, we found that the effect is more reliable for dependencies of subject relations than object relations. Taken together, there is converging evidence from naturalistic corpus data supporting that unpredictable antecedents are encoded with enhanced representation to be more resistant against memory decay and interference, a pattern that is predicted by our SRA.



\newpage

\appendix

\section{Mathematical derivation of strategic resource allocation}

\setcounter{figure}{0}
\renewcommand{\thefigure}{A\arabic{figure}}

\renewcommand{\theequation}{A.\arabic{equation}}
\setcounter{equation}{0}

In this section, we demonstrate the mathematical derivation of strategic resource allocation. First, we will characterize the memory retrieval process as Bayesian inference for the encoded linguistic input (e.g., words). For the purpose of this derivation, we will assume that the linguistic prediction and the underlying memory representation follow Gaussian distributions; these may be interpreted as distributions over values of features. Needless to say, this is a highly simplified view of mental lexicon, and is not necessarily the reality for memory encoding, especially given that word inputs are discrete units rather than continuous variables. However, Gaussian distribution has some desirable mathematical properties with analytical solutions to help us validate the intuition behind our proposal.

\subsection{Memory retrieval via Bayesian inference} \label{sec:retrieval-bayesian}

Suppose one is trying to encode an input word $w$ in noisy memory. We model noise by assuming that the input representation $w$ is corrupted by Gaussian noise with an adjustable precision $\tau_w$, yielding a noisy memory representation $r$:
\begin{equation}
r \sim w + \mathcal{N}\left(0, \tau_w^{-1}\right).
\end{equation}
Retrieval from memory is then performed by forming a reconstructed representation $\hat{w} = \mathop\mathbb{E}_{p(w \mid r)}\left[w\right]$ as the posterior mean on input representations $w$ given noisy memory representations $r$ and a prior distribution on inputs $p_0$, with posterior distribution
\begin{equation}
p(\hat{w} \mid r) \propto p_M(r \mid w) p_0(w).
\end{equation}

Now for ease of analysis, we set the prior distribution on input representations $w$ to be a Gaussian distribution parameterized with mean $w_0$ and the precision $\tau_0$:
\begin{equation}
p_0(w) = \mathcal{N}\left(w \mid w_0, \tau_0^{-1}\right).
\end{equation}
A useful property of Gaussian distribution is that it forms a conjugate prior, so the posterior distribution on inputs words $w$ given memory representations $r$ is also Gaussian distribution. We assume that, in memory retrieval, the decoder   \footnote{This assumes that the decoder has access to the encoding precision $\tau_w$ for the input.}
\begin{equation}
\hat{w} \mid r \sim \mathcal{N}\left(\mu_{\mathrm{post}}, \tau_{\mathrm{post}}^{-1} \right),
\end{equation}
where the posterior mean and precision are
\begin{equation}
\mu_{\mathrm{post}} = \left(1-\alpha_w\right) w_0 + \alpha_w r,\text{    } \tau_{\mathrm{post}} = \tau_0 + \tau_w,
\end{equation}
where $\alpha_w = \frac{\tau_w}{\tau_0 + \tau_w}$.
Then marginalizing out the memory representations, the distribution on reconstructed words is
\begin{equation}
\label{eq:posterior-mean-dist}
\hat{w} \mid w \sim \mathcal{N}\left(\alpha_w w + \left(1-\alpha_w\right) w_0, \frac{\alpha_w}{\tau_0 + \tau_w} \right).
\end{equation}
We see that the retrieved word $\hat{w}$ is pulled towards the prior mode $w_0$ with a weight that depends on the encoding precision $\tau_w$. As the encoding precision $\tau_w$ increases, this attraction to the prior is reduced.


\paragraph{Expected retrieval error under the memory model.}
We define the expected retrieval error $\varepsilon(w)$ for input $w$ as the mean squared error between input $w$ and reconstruction $\hat{w}$:
\begin{equation}
\varepsilon(w) = \mathop\mathbb{E}\left[ \left( \hat{w} - w\right)^2 \mid w \right].
\end{equation}
We can express this mean squared error in terms of the bias--variance decomposition as
\begin{equation}
\varepsilon(w) = \mathrm{Var}[\hat{w} \mid w] + \left(\mathop\mathbb{E}\left[\hat{w} \mid w\right] - w\right)^2.
\end{equation}
Dropping in the mean and variance from Eq.~\ref{eq:posterior-mean-dist}, we can express the bias and variance as
\begin{align}
\mathrm{Var}\left[\hat{w} \mid w\right] &= \frac{\alpha_w}{\tau_0 + \tau_w} \\
\mathop\mathbb{E}\left[\hat{w} \mid w\right] - w &= \alpha_w w + \left(1-\alpha_w\right) w_0 - w \\
&= \left(1-\alpha_w\right) \left(w_0 - w\right).
\end{align}
This gives us a convenient expression for the expected retrieval error,
\begin{align}
\varepsilon(w) &=  \frac{\tau_w}{\left(\tau_w + \tau_0\right)^2} + \left(\frac{\tau_0}{\tau_0 + \tau_w}\left(w_0 - w\right)\right)^2 \\
\label{eq:error-compact}
&= \frac{\tau_w + \tau_0^2\left(w - w_0\right)^2}{\left(\tau_0 + \tau_w\right)^2}.
\end{align}

Furthermore, we will wish to express the expected retrieval error in terms of the surprisal $h_w = -\ln p_0(w)$ of input $w$ and the encoding precision $\tau_w$. From the assumed Gaussian form of the prior over input words, we have
\begin{align}
p_0(w) &= \sqrt{\frac{\tau_0}{2\pi}} \exp\left(-\frac{\tau_0 (w - w_0)^2}{2}\right) \\
h_w &= \frac{\tau_0 (w - w_0)^2}{2} - \ln \sqrt{\frac{\tau_0}{2\pi}}.
\end{align}
Extracting $\tau_0 (w - w_0)^2$ to the left-hand side yields:
\begin{equation}\label{eq:distance}
\tau_0 (w - w_0)^2 = 2 h_w + \ln \frac{\tau_0}{2\pi}.
\end{equation}
Substituting this back into the expression for expected retrieval error, we get
\begin{equation}
\label{eq:expected-error-given-surprisal}
\varepsilon(h_w, \tau_w) = \frac{\tau_w + 2\tau_0 h_w + \tau_0 \ln\frac{\tau_0}{2\pi}}{\left(\tau_0 + \tau_w\right)^2}.
\end{equation}

Below, we will consider how to choose encoding precisions in order to minimize the expected retrieval error under constraints.

\subsection{Minimizing expected error under memory constraint}\label{sec:minimize-error}

In this section, we aim to show that strategic resource allocation arises as a solution to the problem of minimizing the expected retrieval error on average:
\begin{itemize}
    \item \textbf{Strategic Resource Allocation}\\
    Given two linguistic inputs, the minimization of their total expected error bounded by certain memory constraint requires that the more surprising input be encoded with higher precision.
\end{itemize}

Consider two input words $w_1$ and $w_2$ to be encoded and retrieved, whose surprisals are $h_{w_1} = -\ln p_0(w_1)$ and $h_{w_2} = -\ln p_0(w_2)$ respectively. We assume that the encoding precision $\tau_w$ for each word is proportional to the memory resources allocated, and we assume that there is a constraint on total memory resources $c$ allocated for both words, which is to be distributed between $w_1$ and $w_2$. That is, we posit a constraint on the the sum of encoding precisions $\tau_{w_1} + \tau_{w_2} = c$.

We will show that this optimization problem bounded by memory constraint leads to strategic resource allocation in memory encoding, as stated in the Proposition below:

\newtheorem*{proposition*}{Proposition}
\begin{proposition*}
To minimize the total expected retrieval error for two linguistic inputs $\varepsilon_{w_1} + \varepsilon_{w_2}$ subject to a resource constraint $\tau_{w_1} + \tau_{w_2} = c$, the input that is more surprising under the prior distribution must be encoded with higher precision. Specifically, if
\begin{equation}
h_{w_1} > h_{w_2},
\end{equation}
then the optimal encoding satisfies
\begin{equation}
\tau_{w_1} > \tau_{w_2}.
\end{equation}
\end{proposition*}

\begin{proof}

First, in order to minimize expected error $\varepsilon$, we take the derivative of $\varepsilon$ with respect to encoding precision $\tau_w$ for each input\footnote{For simplicity, we maintained $(w - w_0)^2$ for now in Eq.~\ref{eq:derivation-of-epsilon} instead of having it transformed to the form that contains the surprisal $h_w$ of the input.}:
\begin{equation}
\label{eq:derivation-of-epsilon}
\frac{\partial \epsilon(\tau_{w}, h_w)}{\partial \tau_{w}} = \frac{\tau_0 - \tau_w - 2 \tau_0^2 \left( w - w_0 \right)^2}{\left ( \tau_0 + \tau_w \right)^3},
\end{equation}
which reveals three possible situations with respect to the monotonicity of $\varepsilon$ for both inputs:
\begin{enumerate}
    \item The expected error monotonically decreases with increasing $\tau_w$ within its meaningful domain (i.e., $\tau_w > 0$) for both inputs $w_1$ and $w_2$. As shown below, this is the situation for most cases where the input $w$ is not too close to the prior prediction $w_0$ and the prior precision $\tau_0$ is not too unreliable. 
    \item The expected error is a non-monotonic function of $\tau_w$ for both inputs $w_1$ and $w_2$.
    \item The expected error monotonically decreases with $\tau_w$ within in meaningful domain for one input but is non-monotonic for the other.
\end{enumerate}
In this proof, we will show that the proposition above holds in all these three situations.

\subparagraph{Situation 1.} In this first situation, the expected error $\varepsilon$ monotonically decreases for both inputs, which means that
\begin{equation}
\frac{\partial \epsilon(\tau_{w}, h_w)}{\partial \tau_{w}} \leq 0.
\end{equation}
Thus, Situation 1 holds when
\begin{equation}
\tau_0 - \tau_w - 2 \tau_0^2 \left( w - w_0 \right)^2 \leq 0
\end{equation}
\begin{equation}\label{eq:tau-w-inequality}
\tau_w \geq \tau_0 - 2 \tau_0^2 \left( w - w_0 \right)^2.
\end{equation}
Since $\tau_w > 0$, the inequality in Eq.~\ref{eq:tau-w-inequality} always hold within the meaningful domain of $\tau_w$ if 
\begin{equation}
\tau_0 - 2 \tau_0^2 \left( w - w_0 \right)^2 \leq 0
\end{equation}
\begin{equation}
\tau_0 \geq \frac{1}{2 \left( w - w_0 \right)^2}.
\end{equation}
Intuitively, this means that the expected error $\varepsilon$ monotonically decreases if the input $w$ is not too close to the prior prediction $w_0$ and the prior precision $\tau_0$ is not too unreliable.

Now that $\varepsilon$ monotonically decreases within the meaningful domain of $\tau_w$ for both inputs $w_1$ and $w_2$ as defined in this first situation, using Eq.~\ref{eq:expected-error-given-surprisal}, we evaluate the derivative with respect to its relationship with input surprisal
\begin{equation}\label{eq:error-derivative-w-surp}
\begin{split}
f(\tau_w, h_w) &= \frac{\partial \epsilon(\tau_{w}, h_w)}{\partial \tau_{w}} \\
&= \frac{\tau_0 - \tau_w - 2\tau_0 \left(2 h(w) + \ln \left( \frac{\tau_0}{2\pi} \right) \right)}{(\tau_0 + \tau_w)^3}\\
&< 0\text{ (by the definition of Situation 1)},
\end{split}
\end{equation}
which shows that $f(\tau_w, h_w)$ monotonically decreases in $h_w$. 

When $h_{w_1} = h_{w_2}$, $f(\tau_{w_1}, h_{w_1})$ and $f(\tau_{w_2}, h_{w_2})$ have the same quantitative form. As a result, the memory resources will be uniformly distributed across $w_1$ and $w_2$ with no impetus to redistribute more resources to any one of them:
\begin{equation}
\label{eq:uniform-distribute}
\tau_{w_1} = \tau_{w_2} = \frac{c}{2}.
\end{equation}

However, when $h_{w_1} > h_{w_2}$, since $f(\tau_w, h_w)$ monotonically decreases in $h_w$, we have $f(\tau_{w_1}, h_{w_1})$ < $f(\tau_{w_2}, h_{w_2})$ at any given value of $\tau_w$. As a result, compared to the uniform distribution in Eq.~\ref{eq:uniform-distribute}, there is reason to redistribute more resources to to the high surprisal $w_1$, since the decrease of error on $w_1$ will be higher than the increase of error on $w_2$, yielding a lower total error across the two inputs. Therefore, in Situation~1, if $h_{w_1} > h_{w_2}$, the optimal encoding strategy should satisfy $\tau_{w_1} > \tau_{w_2}$.

\begin{figure}
    \centering
    \includegraphics[width=0.4\linewidth]{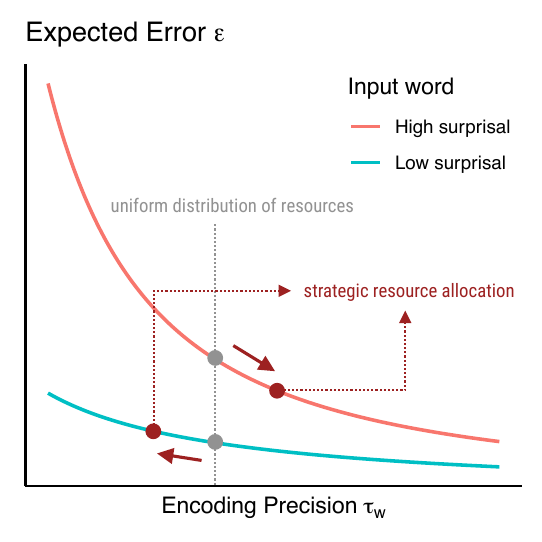}
    \caption{Situation~1 expected retrieval error $\epsilon$ for two input words as a function of encoding precision $\tau_w$ and their surprisal.}
    \label{fig:func-error}
\end{figure}

\subparagraph{Situation 2.} As shown above, in order for $\varepsilon$ to be a non-monotonic function of $\tau_w$:
\begin{equation}\label{eq:monotonicity-turning-point}
\tau_0 < \frac{1}{2 \left( w - w_0 \right)^2},
\end{equation}
which corresponds to borderline cases where the input $w$ is too close to the prior prediction $w_0$ or the prior precision $\tau_0$ is too unreliable. 

In this second situation, with increasing $\tau_w$, $\varepsilon$ first increases and then decreases, as illustrated in Figure~\ref{fig:func-error-nonmono}. And the relationship between high surprisal and low surprisal inputs has three phases. We will show that the proposition still holds for all the three phases in Situation 2, such that the optimal encoding satisfies $\tau_{w_1} > \tau_{w_2}$ if $h_{w_1} > h_{w_2}$.
\begin{figure}
    \centering
    \includegraphics[width=0.4\linewidth]{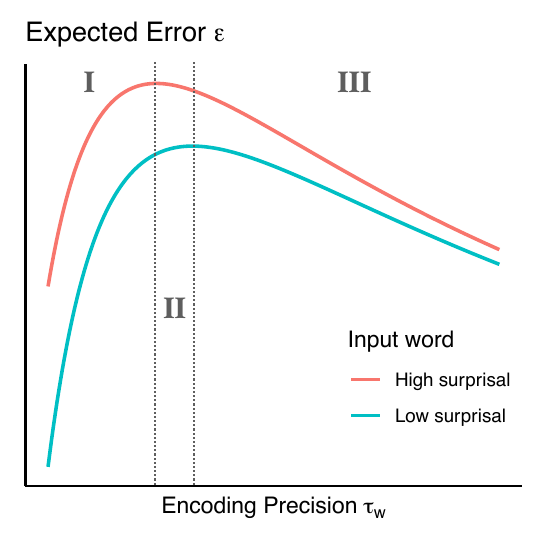}
    \caption{Situation~2 expected retrieval error $\epsilon$ for two input words as a function of encoding precision $\tau_w$ and their surprisal.}
    \label{fig:func-error-nonmono}
\end{figure}

In \textbf{Phase~III}, the situation is basically the same as the Situation~1 discussed above, where $\varepsilon$ decreases with increasing $\tau_w$ for both inputs. Therefore, according to the proof in Situation~1, the optimal encoding satisfies $\tau_{w_1} > \tau_{w_2}$ if $h_{w_1} > h_{w_2}$.

In \textbf{Phase~II}, $\varepsilon$ decreases in $\tau_w$ for one input and increases for the other. For each input, the turning point where the monotonicity is flipped is at
\begin{equation}
\tau_w = \tau_0 - 2 \tau_0 \left( 2 h(w) + \ln \left(\frac{\tau_0}{2\pi} \right) \right),
\end{equation}
where the derivative of $\varepsilon$ in Eq.~\ref{eq:error-derivative-w-surp} is 0. Importantly, if $h_{w_1} > h_{w_2}$, then the turning points $\tau_{w_1} < \tau_{w_2}$. Therefore, the turning point for the high surprisal input $w_1$ is to the left of the one for the low surprisal input $w_2$. That means, in Phase~II, it can only be the case that the expected error $\varepsilon$ decreases in $\tau_w$ for the high surprisal $w_1$, but increases for the low surprisal $w_2$. As a result, in order to minimize $\varepsilon$, more memory resources should be allocated to encode $w_1$ than $w_2$, leading to $\tau_{w_1} > \tau_{w_2}$.

In \textbf{Phase~I}, $\varepsilon$ increases in $\tau_w$ for both inputs $w_1$ and $w_2$. Recall that the derivative of $\varepsilon$ (repeated below in Eq.~\ref{eq:error-derivative-w-surp-2}) decreases as the input surprisal $h_w$ increases. 
\begin{equation}\label{eq:error-derivative-w-surp-2}
\begin{split}
f(\tau_w, h_w) &= \frac{\partial \epsilon(\tau_{w}, h_w)}{\partial \tau_{w}} \\
&= \frac{\tau_0 - \tau_w}{(\tau_0 + \tau_w)^3} - \frac{2\tau_0 \left(2 h(w) + \ln \left( \frac{\tau_0}{2\pi} \right) \right)}{(\tau_0 + \tau_w)^3}\\
&> 0.
\end{split}
\end{equation}
Therefore, in Phase~I, the expected error $\varepsilon$ increases more slowly for high surprisal input. As a result, if $h_{w_1} > h_{w_2}$, then for a fixed amount of memory resources, more resources allocated to $w_1$ yields lower smaller increase in $\varepsilon$, leading to $\tau_{w_1} > \tau_{w_2}$.

\subparagraph{Situation 3.} In this third situation, the expected error $\varepsilon$ monotonically decreases in $\tau_w$ for one input, but is a non-monotonic function for the other.
\begin{figure}
    \centering
    \includegraphics[width=0.4\linewidth]{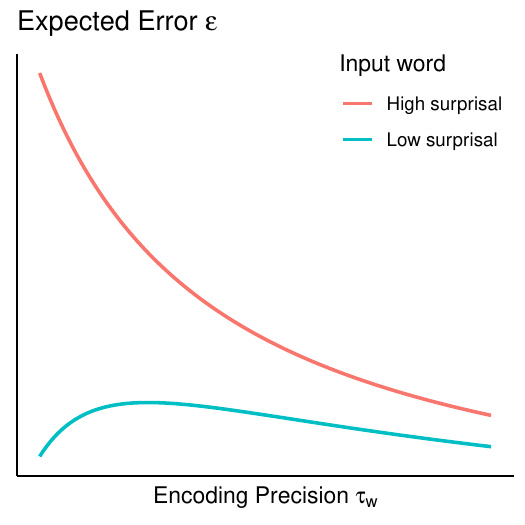}
    \caption{Situation~3 expected retrieval error $\epsilon$ for two input words as a function of encoding precision $\tau_w$ and their surprisal.}
    \label{fig:func-error-divergent}
\end{figure}

Recall that whether $\varepsilon$ is monotonic depends on the inequality in Eq.~\ref{eq:monotonicity-turning-point}. That is, in order for $\varepsilon$ to be non-monotonic, Eq.~\ref{eq:monotonicity-turning-point} must hold. Moreover, as discussed in Situation~2 above, if $h_{w_1} > h_{w_2}$, the turning point of monotonicity for the high surprisal input $w_1$ is to the left of the one for the low surprisal input $w_2$. As a result, in this Situation~3, it must be the case that it is the high surprisal input $w_1$ that monotonically decreases in $\tau_w$ whereas the low surprisal input $w_2$ first increases and then decreases, as illustrated in Figure~\ref{fig:func-error-divergent}.

Apparently, Situation~3 is basically equivalent to the Phase~II and Phase~III in Situation~2. Therefore, as proved above, if $h_{w_1} > h_{w_2}$, the optimal encoding should satisfy $\tau_{w_1} > \tau_{w_2}$. 

\end{proof}

\paragraph{Remark.} To sum up, if the surprisal of two input words $h_{w_1} > h_{w_2}$, given fixed amount of memory resources such that the encoding precisions for two inputs is constrained by $\tau_{w_1} + \tau_{w_2} = c$, the optimal encoding with strategic resource allocation should satisfy $\tau_{w_1} > \tau_{w_2}$ in order to achieve minimal total expected error $\varepsilon$. We outlined three possible situations of how $\varepsilon$ may change with increasing encoding precision $\tau_w$, and we proved that the strategic resource allocation should hold in all three situations. It is worth noting that, in most cases, $\varepsilon$ monotonically decreases with increasing encoding precision $\tau_w$, as in Situation~1. However, when the prior precision is too unreliable or when the input word is too close to the prior predicted word, there will be borderline cases where $\varepsilon$ first increases with increasing $\tau_w$ before it starts to decrease, as in Situation~2 and 3.

\section{Statistical models}\label{sec:stats-models}

\newcounter{list-counter-AppendixB}

\subsection{Study 1}

In Study~1, for each language, we ran regression models on dependency length $L$. As mentioned in the main article, the regression models were run separately for $L$ measured as intervening word counts and as intervening surprisal, as shown below in (\ref{model-study1-word}) and (\ref{model-study1-surp}). For the analysis on full dataset, we ran linear mixed-effect model with random intercept per dependency type. For analysis on subject relations and object relations, we ran the standard linear regression without specifying random effects. Compare to the orthographic $L$, the analysis with information-theoretic $L$ includes \textsc{baseline surprisal} as an additional control variable.

\begin{enumerate}[(1)]
\setcounter{enumi}{\value{list-counter-AppendixB}}
    \item \label{model-study1-word} Regression formulas for orthographic $L_{\mathrm{O}}$
    \begin{itemize}
        \item \textit{Full dataset} \\
             $ L \sim$ 1 + Sentence Position + Antecedent Position + Sentence Length + Antecedent Frequency + Antecedent Surprisal + (1 $\mid$ Dependency Type)
        \item \textit{Subject/object relations} \\
            $ L \sim$ 1 + Sentence Position + Antecedent Position + Sentence Length + Antecedent Frequency + Antecedent Surprisal
    \end{itemize}
\setcounter{list-counter-AppendixB}{\value{enumi}}
\end{enumerate}
\begin{enumerate}[(1)]
\setcounter{enumi}{\value{list-counter-AppendixB}}
    \item \label{model-study1-surp} Regression formulas for information-theoretic $L_{\mathrm{I}}$
    \begin{itemize}
        \item \textit{Full dataset} \\
             $ L \sim$ 1 + Sentence Position + Antecedent Position + Sentence Length + Baseline Surprisal + Antecedent Frequency + Antecedent Surprisal + (1 $\mid$ Dependency Type)
        \item \textit{Subject/object relations} \\
            $ L \sim$ 1 + Sentence Position + Antecedent Position + Sentence Length + Baseline Surprisal + Antecedent Frequency + Antecedent Surprisal
    \end{itemize}
\setcounter{list-counter-AppendixB}{\value{enumi}}
\end{enumerate}

\subsection{Study 2a}

In Study~2a, we ran linear mixed-effect models on log-transformed reading times for the critical region at the retrieval site (i.e., the right codependent for each syntactic dependency) and its spillover region.

\begin{enumerate}[(1)]
\setcounter{enumi}{\value{list-counter-AppendixB}}
    \item \label{model-study2a-critical} Regression formulas for the critical region
    \begin{enumerate}[a.]
        \item Fixed effects \\
            logRT $\sim$ 1 + sent.pos + antec.pos + sent.len + word.len + antec.freq +
             surp + surp.prev1 + surp.prev2 +
             freq + freq.prev1 + freq.prev2 +
             $L$ * antec.surp
        \item Random effects
        \begin{itemize}
            \item Orthographic $L_\mathrm{O}$
            \begin{itemize}
                \item \textit{Full dataset}: (1 $\mid$ dep.type) + (1 $\mid$ part)
                \item \textit{Subject relations}: ($L$ + antec.surp $\mid$ part)
                \item \textit{Object relations}: ($L$ + antec.surp $\mid$ part)
            \end{itemize}
            \item Info-theoretic $L_\mathrm{I}$
            \begin{itemize}
                \item \textit{Full dataset}: (1 $\mid$ dep.type) + (1 $\mid$ part)
                \item \textit{Subject relations}: (antec.surp $\mid$ part)
                \item \textit{Object relations}: ($L$ + antec.surp $\mid$ part)
            \end{itemize}
        \end{itemize}
    \end{enumerate}
\setcounter{list-counter-AppendixB}{\value{enumi}}
\end{enumerate}

\begin{enumerate}[(1)]
\setcounter{enumi}{\value{list-counter-AppendixB}}
    \item \label{model-study2a-critical} Regression formulas for the spillover region
    \begin{enumerate}[a.]
        \item Fixed effects \\
            logRT $\sim$ 1 + sent.pos + antec.pos + sent.len + word.len + antec.freq +
             surp + surp.prev1 + surp.prev2 +
             freq + freq.prev1 + freq.prev2 +
             $L$ * antec.surp
        \item Random effects
        \begin{itemize}
            \item Orthographic $L_\mathrm{O}$
            \begin{itemize}
                \item \textit{Full dataset}: (1 $\mid$ dep.type) + (1 $\mid$ part)
                \item \textit{Subject relations}: ($L$ * antec.surp $\mid$ part)
                \item \textit{Object relations}: (antec.surp $\mid$ part)
            \end{itemize}
            \item Info-theoretic $L_\mathrm{I}$
            \begin{itemize}
                \item \textit{Full dataset}: (1 $\mid$ dep.type) + (1 $\mid$ part)
                \item \textit{Subject relations}: (antec.surp $\mid$ part)
                \item \textit{Object relations}: (antec.surp $\mid$ part)
            \end{itemize}
        \end{itemize}
    \end{enumerate}
\setcounter{list-counter-AppendixB}{\value{enumi}}
\end{enumerate}

\subsection{Study 2b}

In Study~2b, we ran linear mixed-effect models on first-pass durations and total reading times at the retrieval site, as shown below in (\ref{model-study2b-first-pass}) and (\ref{model-study2b-total-RT}).

\begin{enumerate}[(1)]
\setcounter{enumi}{\value{list-counter-AppendixB}}
    \item \label{model-study2b-first-pass} Regression formulas for first-pass durations (same maximal converging random structure for both measures of $L$)
    \begin{enumerate}[a.]
        \item Fixed effects \\
            logRT $\sim$ 1 + sent.pos + antec.pos + sent.len + word.len + antec.freq +
             surp + surp.prev1 + surp.prev2 +
             freq + freq.prev1 + freq.prev2 +
             $L$ * antec.surp
        \item Random effects
        \begin{itemize}
            \item \textit{Full dataset}: ($L$ * antec.surp $\mid$ dep.type) + ($L$ * antec.surp $\mid$ part)
            \item \textit{Subject relations}: ($L$ * antec.surp $\mid$ part)
            \item \textit{Object relations}: ($L$ * antec.surp $\mid$ part)
        \end{itemize}
    \end{enumerate}
\setcounter{list-counter-AppendixB}{\value{enumi}}
\end{enumerate}

\begin{enumerate}[(1)]
\setcounter{enumi}{\value{list-counter-AppendixB}}
    \item \label{model-study2b-total-RT} Regression formulas for total reading times (same maximal converging random structure for both measures of $L$ in the analysis of full dataset and subject relations)
    \begin{enumerate}[a.]
        \item Fixed effects \\
            logRT $\sim$ 1 + sent.pos + antec.pos + sent.len + word.len + antec.freq +
             surp + surp.prev1 + surp.prev2 +
             freq + freq.prev1 + freq.prev2 +
             $L$ * antec.surp
        \item Random effects
        \begin{itemize}
            \item \textit{Full dataset}: ($L$ * antec.surp $\mid$ dep.type) + ($L$ * antec.surp $\mid$ part)
            \item \textit{Subject relations}: ($L$ * antec.surp $\mid$ part)
            \item \textit{Object relations}: 
            \begin{itemize}
                \item Orthographic $L_\mathrm{O}$: ($L$ * antec.surp $\mid$ part)
                \item Info-theoretic $L_\mathrm{I}$: ($L$ + antec.surp $\mid$ part)
            \end{itemize}
        \end{itemize}
    \end{enumerate}
\setcounter{list-counter-AppendixB}{\value{enumi}}
\end{enumerate}

\newpage

\medskip

\bibliography{weijie_references.bib} 

\newpage

\end{document}